\documentclass[10pt,twocolumn,letterpaper]{article}

\usepackage{iccv}
\usepackage{times}
\usepackage{epsfig}
\usepackage{graphicx}
\usepackage{amsmath}
\usepackage{amssymb}
\usepackage[dvipsnames]{xcolor}
\usepackage{colortbl}
\usepackage[font=small]{caption}
\usepackage[font=small]{subcaption}
\usepackage{algorithmic}
\usepackage[ruled, linesnumbered]{algorithm2e}
\usepackage{multirow, multicol}
\usepackage{tabularray}
\usepackage[accsupp]{axessibility}  

\usepackage[pagebackref=true,breaklinks=true,letterpaper=true,colorlinks,bookmarks=false]{hyperref}

\iccvfinalcopy 

\def\iccvPaperID{2874} 

\newcommand{\supp}[1]{{\color{blue}  #1}}
\newcommand{\main}[1]{{\color{blue}  #1}}
\begin{document}

\title{TeD-SPAD: Temporal Distinctiveness for Self-supervised Privacy-preservation for video Anomaly Detection}

\author{Joseph Fioresi, Ishan Rajendrakumar Dave, Mubarak Shah\\
Center for Research in Computer Vision,
University of Central Florida, Orlando, USA\\
{\tt\small \{joseph.fioresi, ishanrajendrakumar.dave\}@ucf.edu, shah@crcv.ucf.edu}\\
{\tt\small Project Page: \url{https://joefioresi718.github.io/TeD-SPAD_webpage/}}
}
\maketitle

\begin{abstract}
    Video anomaly detection (VAD) without human monitoring is a complex computer vision task that can have a positive impact on society if implemented successfully. While recent advances have made significant progress in solving this task, most existing approaches overlook a critical real-world concern: privacy. With the increasing popularity of artificial intelligence technologies, it becomes crucial to implement proper AI ethics into their development. Privacy leakage in VAD allows models to pick up and amplify unnecessary biases related to people's personal information, which may lead to undesirable decision making. In this paper, we propose TeD-SPAD, a privacy-aware video anomaly detection framework that destroys visual private information in a self-supervised manner. In particular, we propose the use of a temporally-distinct triplet loss to promote temporally discriminative features, which complements current weakly-supervised VAD methods. Using TeD-SPAD, we achieve a positive trade-off between privacy protection and utility anomaly detection performance on three popular weakly supervised VAD datasets: UCF-Crime, XD-Violence, and ShanghaiTech. Our proposed anonymization model reduces private attribute prediction by 32.25\% while only reducing frame-level ROC AUC on the UCF-Crime anomaly detection dataset by 3.69\%.
\end{abstract}
\vspace{-2mm}

\section{Introduction}

Machine learning-driven technologies are increasingly being adopted by society. The progress in cloud computing has enabled the deployment of even computationally intensive technologies in the public space. One such application is video anomaly detection (VAD) in autonomous video analytics. VAD is a video understanding task that aims to identify the temporal location of anomalous events occurring in long continuous videos without human supervision. An anomaly can be defined as any unusual event, such as a traffic accident, an elderly person falling, or a fire.
Proper application of this technology can result in faster response times to anomalies, without the need for human resources to monitor camera feeds.

\begin{figure}
    \centering
    \includegraphics[width=\linewidth]{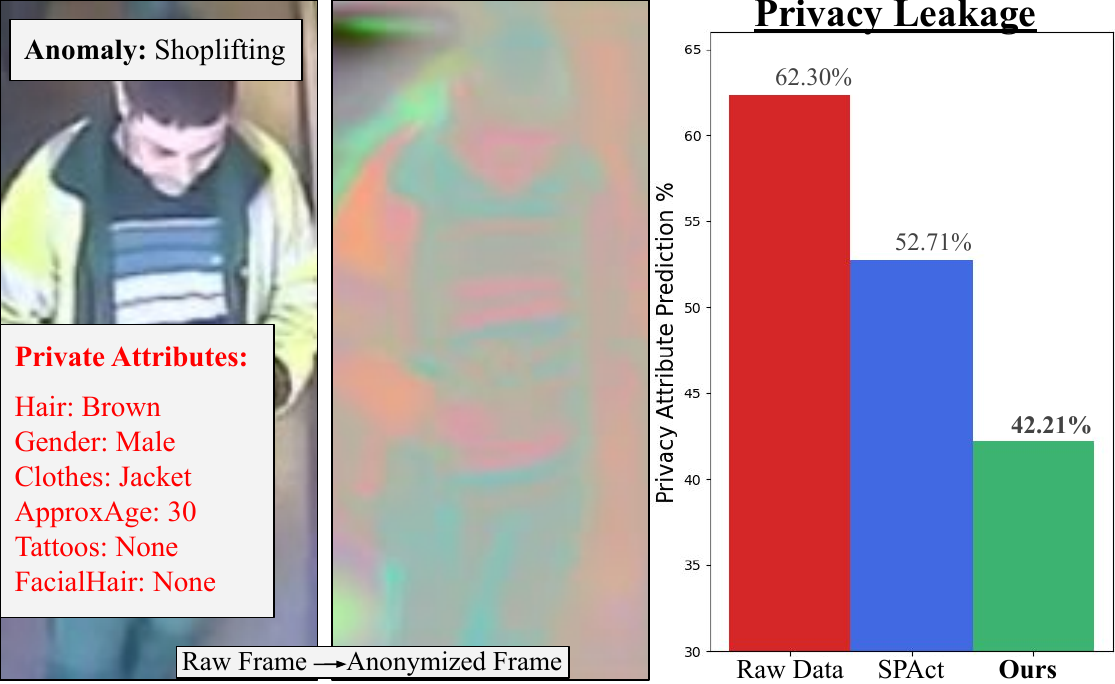}
    \caption{Single frame from video Shoplifting052\_x264.mp4 of UCF-Crime \cite{sultani_real-world_2018} showing the types of private attributes (shown in red on the left image) leaked in visual data. In the right image we show the anonymized frame where these attributes are barely visible. The graph demonstrates the ability of our self-supervised framework to mitigate privacy leakage compared to prior work SPAct~\cite{spact}. At a similar anomaly detection utility performance, our method prevents \textbf{32\%} of the visual private leakage compared to raw data.}
    \label{fig:teaser}
\end{figure}

However, public adoption of such AI technologies brings justifiable concern about safety and their decision-making capabilities. Many of these concerns center around privacy violations and accuracy. VAD is an application where visual \textit{privacy leakage} and \textit{data bias} are exceedingly important issues. Sending videos to remote computers or cloud services to process results in unnecessary privacy leakage for people who are not directly involved in anomalous events. Additionally, an application employing a standard RGB video will incorporate any bias found in its training set, leading to potentially unfair decisions. An illustration of privacy leakage is shown in Fig.~\ref{fig:teaser}.
 
Recently, there have been interesting attempts to prevent visual privacy leakage in action recognition. Some of the approaches utilize input downsizing-based solutions~\cite{ryoo, ishwar, liu2020indoor} and object-detection-dependent obfuscation-based formulations~\cite{ren2018learning, zhang2021multi}. Wu~\etal~\cite{wang2019privacy} proposed an adversarial training-based framework where they train an anonymization function to remove privacy-preservation. Dave~\etal~\cite{spact} proposed a self-supervised privacy-preserving framework that does not require privacy labels and achieves similar performance to the supervised method~\cite{wang2019privacy}.
Since many weakly-supervised anomaly detection (WSAD) methods rely on the pretrained features of action recognition, privacy-preserving action recognition seems like a promising candidate for privacy-preserving anomaly detection. However, detecting anomalies does not align well with privacy-preserved action recognition videos. The use of short videos in action recognition encourages the use of \textit{temporally-invariant} features, where the features of clips at distinct timesteps should be the same. Conversely, detecting anomalies in long, untrimmed videos requires \textit{temporally-distinct} reasoning, where features of clips at distinct timesteps of a video should be different, to determine whether events in the \textit{same scene} are anomalous. This is why most existing anomaly detection methods focus on refining the features of pretrained video encoders to increase their temporal separability.

To the best of our knowledge, privacy-preservation in video anomaly detection is an unexplored area in computer vision. Building on the existing self-supervised privacy-preserving action recognition framework~\cite{spact}, we propose a more aligned utility branch for anomaly detection. To achieve this, we use a novel temporally-distinct triplet loss to promote temporal distinctiveness during anonymization training. Once the anonymization function is learned through our proposed anonymization framework, we apply it to the anomaly dataset, which ensures privacy leakage mitigation in privacy-sensitive anomaly detection tasks. We use these anonymized features to train the current state-of-the-art WSAD method, MGFN~\cite{chen_mgfn_2022}.

To evaluate privacy-preserving performance in anomaly detection, we adopt protocols from prior action recognition methods, where we report the utility performance on WSAD task on widely used anomaly datasets (UCF-Crime~\cite{sultani_real-world_2018}, XD-Violence~\cite{wu2020not}, and ShanghaiTech~\cite{liu2018ano_pred}) and budget performance on privacy dataset VISPR~\cite{orekondy17iccv}.

Our contributions can be summarized as follows:

\begin{itemize}
    \item We introduce a new problem of privacy-preservation in video anomaly detection, where we identify the privacy leakage issue in existing weakly supervised anomaly detection methods.
    \item To address this open problem, we propose TeD-SPAD, a framework based on self-supervised privacy-preservation with a temporally-distinct triplet loss to make the video anonymization process more suitable for anomaly detection. 
    \item We propose evaluation protocols for privacy vs. anomaly trade-off,  demonstrating that our proposed framework outperforms prior methods by significant margins across all anomaly detection benchmarks. On the widely used UCF-Crime dataset, our method is able to eliminate \textbf{32.25\%} of the privacy leakage at a cost of only a 3.96\% reduction in frame-level AUC performance.
\end{itemize}

\section{Related Works}
\noindent\textbf{Privacy Preservation}

We observe that many works preserve visual privacy at capture time by using non-intrusive sensors such as thermal imaging, depth cameras, or event cameras \cite{luo_computer_nodate, hinojosa_privhar_2022, kim_privacy-preserving_2022, Ahmad2023eventreid}. Other works allow for raw RGB visual information to be captured, but make an effort to protect the subject privacy in such a way that the data is still useful in a utility task. Earlier efforts aimed at dealing with visual privacy include image downsampling \cite{dai_towards_2015} or blocking/blurring privacy-related objects located using pretrained object detectors. Both of these obfuscation methods were shown to reduce utility results by more than they reduced privacy leakage \cite{spact, li_stprivacy_2023, wang2019privacy}.

Recent developments have yielded numerous privacy-preserving approaches to action recognition~\cite{ahmad2022event, zehtabian2021privacy, moon2023anonymization}. Wu~\etal released an action dataset with privacy labels, PA-HMDB \cite{wang2019privacy}. They use an adversarial learning framework to obfuscate the privacy features using supervised privacy labels. MaSS \cite{chen2022mass} uses a similar framework to Wu~\etal \cite{wang2019privacy}, except it adapts a compound loss to flexibly preserve certain attributes instead of destroying them. STPrivacy \cite{li_stprivacy_2023} upgraded the general framework to work with a transformer anonymizing block, masking entire video tubelets unnecessary for action recognition. Following Dave~\etal's SPAct \cite{spact}, we adopt a similar self-supervised adversarial anonymization framework without the use of the privacy labels, using NT-Xent~\cite{chen_simple_2020} contrastive loss in the budget branch to mitigate spatial privacy leakage.

\noindent\textbf{Anomaly Detection}

With such a high volume of available video footage, it is infeasible to create sufficient labelled data to solve supervised VAD. Therefore, many works explore unsupervised methods. These generally train a reconstruction model, then either reconstruct the current frame or try to predict the next frame, signaling an anomaly when reconstruction error is high \cite{cong_sparse_2011, popoola_video-based_2012, kiran_overview_2018, park_learning_2020, yuan_transanomaly_2021, sun_unsupervised_2023}. Giorgi~\etal \cite{giorgi_privacy-preserving_2022} used an autoencoder with differential privacy, generating anomaly scores from the noisy reconstructed images. This method helped retain some level of subject privacy, but was only evaluated on image quality metrics.

Sultani~\etal \cite{sultani_real-world_2018} brought weak supervision to VAD, where anomalies are labelled at the video level. These authors introduce UCF-Crime, a large scale weakly supervised dataset. They propose formulating weakly supervised VAD as a multi instance learning (MIL) problem, showcasing the benefits of temporal smoothness loss \& sparsity loss. With the exception of a select few works \cite{zhong_graph_2019, zhang_exploiting_2022}, all following weakly supervised methods are considered anomaly feature representation learning as they improve upon MIL formulation \cite{zhu_motion-aware_2019, tian_weakly-supervised_2021, feng2021mist, li_self-training_2022, huang_weakly_2022, wu_s3r_2022, chen_mgfn_2022, sapkota_bayesian_2022}, which involves interpreting static video features extracted using an action classifier. Zhong~\etal~\cite{zhong_graph_2019} propose a rearranged weakly supervised version of ShanghaiTech. Wu~\etal \cite{wu2020not} introduce XD-Violence, a large-scale weakly supervised dataset that includes audio, bringing multi-modal fusion to VAD. In this work, we choose to focus on the weakly supervised video anomaly detection setting due to its effectiveness and low annotation effort.

Most VAD works find it useful to model temporal relations between video segments \cite{tian_weakly-supervised_2021, feng2021mist, lv_localizing_2021, wu2020not, li_self-training_2022, wu_s3r_2022, zhu_motion-aware_2019, chen_spatialtemporal_2023}. \cite{lv_localizing_2021} are able to exploit dynamic variations in features between consecutive segments to help localize anomalies. 
The authors of RTFM \cite{tian_weakly-supervised_2021} find that anomalous segments tend to have larger feature magnitudes ($\ell_2$ norm of extracted clip feature vector) than normal segments. They introduce a feature magnitude learning function to help identify anomalous segments with large feature magnitudes. MGFN \cite{chen_mgfn_2022} notices that varying scene attributes causes some normal videos to have larger feature magnitudes than anomalous, so they propose a feature magnitude contrastive loss to help capture these instances. In this work, we use the MGFN model to evaluate anomaly detection performance. 
Complementing these ideas, \cite{wu_learning_2021} demonstrates the effectiveness of explicitly encouraging feature discrimination.
Intuitively, these observations can be aggregated by enforcing temporally-distinct feature representations. 
Building on the terminology introduced in ~\cite{patrick2021compositions} regarding invariant and distinctive representations, in this paper our goal is to enhance the distinctiveness of the representations within our utility branch.

\begin{figure*}
    \centering
    \includegraphics[width=0.85\textwidth]{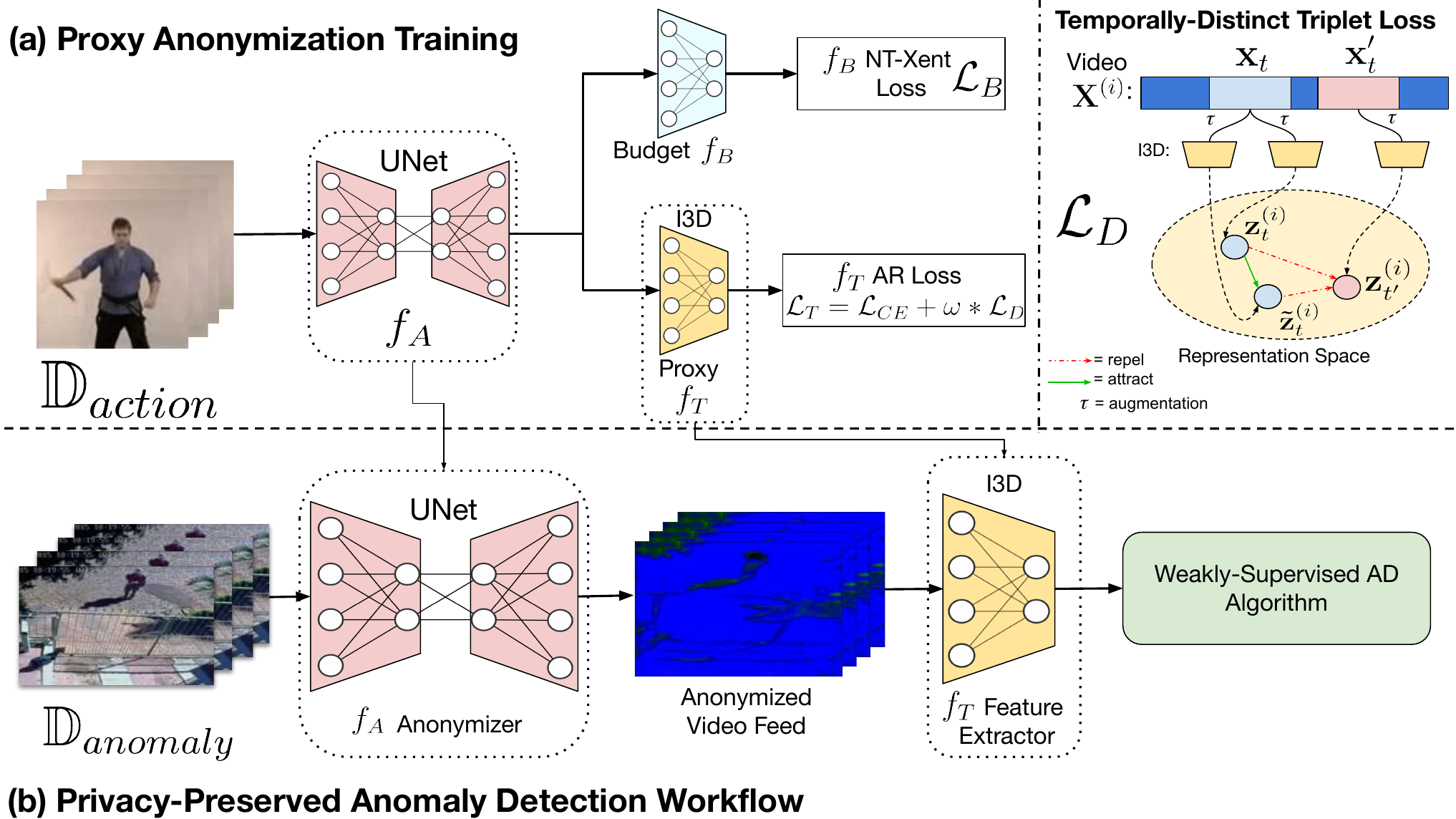}
    \caption{Full TeD-SPAD framework consisting of the proxy anonymization training followed by the privacy-preserved anomaly detection. (a) shows this proxy training, where UNet is used to anonymize frames in such a way that reduces mutual information between frames while maintaining utility performance. We complement the standard cross-entropy loss with our proposed temporally-distinct triplet loss, which enforces a difference in clip features at distinct timesteps. After training the anonymizer and feature extractor, (b) shows the privacy-preserved workflow, where the anomaly dataset videos are passed through the proxy-trained $f_A$, $f_T$, then into any WSAD algorithm.}
    \label{fig:workflow_diagram}
\end{figure*}

\section{Method}

The central idea behind our proposed framework is to develop an anonymization function that can degrade privacy attributes during training without relying on privacy labels. Furthermore, this function must be able to maintain the performance of the weakly-supervised anomaly detection task. Fig.~\ref{fig:workflow_diagram} displays a schematic diagram of the proposed framework. In Sec.~\ref{sec:problem}, we provide a comprehensive discussion of the problem statement. Next, in Sec.~\ref{sec:framework}, we describe the component of the framework and their initialization process. Sec.~\ref{sec:anony} outlines the anonymization function training, where we propose a temporally-distinct triplet loss to enhance the existing self-supervised privacy-preserving framework~\cite{spact}. Once we learn the anonymization function, in Sec.~\ref{sec:mgfn}, we train the anomaly detection model using the privacy-preserved features obtained through our anonymization function. An overview of our complete framework is outlined in Algorithm Sec.~\ref{sec:algo}.

\subsection{Problem Statement}
\label{sec:problem}

We define our problem statement similar to privacy-preserving action recognition frameworks~\cite{spact, wu_learning_2021}, but, with a different utility task. Suppose we have a video dataset $\mathbb{D}_{anomaly}=\{X^{(i)}, Y^i\}_{i=1}^{N}$, where, $X^{(i)}$ is a video-instance, $N$ is the total number of samples and $Y^{(i)} \in \{0,1\}$ is a binary label. Considering video-level anomaly detection as a utility task $T$, and privacy attribute classification as the budget task $B$, the aim of a privacy-preserving system is to maintain the performance of $T$ while reducing $B$. To achieve this goal, the system learns an anonymization function $f_{A}$, which modifies the original raw data. This goal of privacy-preservation could be fundamentally expressed as following criteria:

\noindent\textit{Criterion-1}: The performance of the utility task should not degrade from the original performance, i.e loss $\mathcal{L}_{T}$ value of the utility target model $f'_{T}$ should remain almost identical before and after applying the anonymization function. 
\begin{equation}
    \mathcal{L}_{T}(f'_{T}(f^{*}_{A}(X)), Y) \approx \mathcal{L}_{T}(f'_{T}(X), Y).
    \label{eq:ft_criterion}
\end{equation}

\noindent\textit{Criterion-2}: Applying anonymization function should increase the loss $\mathcal{L}_{B}$ for budget $B$ task of target budget model $f'_{B}$.
\begin{equation}
    \mathcal{L}_{B}(f'_{B}(f^{*}_{A}(X)))  \gg  \mathcal{L}_{B}(f'_{B}(X)).
    \label{eq:fb_criterion}
\end{equation}

Regarding weakly-supervised anomaly detection (WSAD), most of the existing methods require multi-stage training, meaning they are not end-to-end. This presents a challenge for incorporating it as a utility task in anonymization training. By contrast, privacy-preserving action recognition frameworks~\cite{spact,wang2019privacy} have an end-to-end utility task (i.e action recognition), making it more straightforward to include. 

Since most WSAD methods rely on pretrained video encoder features from large-scale action recognition training, we can utilize the exact same anonymization process of privacy-preserving action recognition~\cite{spact}, utilizing action recognition as a \textit{proxy-utility task} on a proxy-utility action dataset, which we notate as ${\mathbb{D}_{action}}$. The limitation of such anonymization training is that by focusing solely on optimizing \textit{short clips} for action recognition, it neglects the importance of temporally-distinct features. This oversight leads to a notable decline in anomaly detection performance, which relies heavily on these temporally-distinct features, as the training progresses. To resolve this issue, we reformulate the anonymization utility branch to enforce temporal distinctiveness to better align with the anomaly detection task. Hence, our $\mathcal{L}_{T}$ is weighted sum of action recognition loss ($\mathcal{L}_{CE}$) and temporally-distinct triplet loss ($\mathcal{L}_{D}$) which is elaborated in Eq.~\ref{eq:utility}.

\subsection{Anonymization Framework} 
\label{sec:framework}
Our anonymization framework consists of 3 main components: (1) Anonymization function ($f_A$), which is a simple encoder-decoder model with a sigmoid activation. (2) Privacy removal model ($f_B$), which is an image encoder.  (3) Utility model ($f_T$), which is a video encoder. 

\noindent \textbf{Network Initialization} First of all, our anonymization model is pretrained to initialize as an identity function. This pretraining involves the reconstruction of the frames from $\mathbb{D}_{action}$ using L1-reconstruction loss. 

\begin{equation}
    \mathcal{L}_{L1} = \sum_{c=1}^{C}\sum_{h=1}^{H}\sum_{w=1}^{W}|x_{c,h,w}- \hat{x}_{c,h,w}|,
    \label{eq:l1loss}
\end{equation}
where $\hat{x}$ is the output of $f_A$, $x$ is an input image, and $C,H,W$ corresponds to the channel, height, and width of the input image. 

Secondly, the privacy model $f_B$ is initialized with the self-supervised weights of SimCLR~\cite{simclr} on ImageNet~\cite{deng2009imagenet}. The video encoder $f_T$ is pretrained with the standard action recognition weights from Kinetics400 dataset~\cite{carreira_quo_2018}.

\subsection{Anonymization Training}
Anonymization training is mainly comprised of a minimax optimization of utility loss and privacy loss. 
\label{sec:anony}

\noindent{\bf Temporally-distinct triplet loss as Utility}
Weakly-supervised video anomaly detection methods leverage temporal information to help localize anomalies. \cite{lv_localizing_2021, huang_weakly_2022, tian_weakly-supervised_2021, chen_mgfn_2022} show the positive effect of feature magnitude ($\ell_2$ norm of clip feature vector) separability along the temporal dimension. In order to adopt the SPAct anonymization framework to anomaly detection problems, we utilize a temporal distinctiveness objective in the utility branch. We design a temporally-distinct triplet loss, which increases the agreement between temporally-aligned clips of the same video and increases dissimilarity between representations of the clips which are temporally-misaligned. For anchor clip $\mathbf{x}^{(i)}_{t}$, we obtain the positive clip from the exact same timestamp, but with a differently augmented version denoted as $\mathbf{\tilde{x}}^{(i)}_{t}$. Whereas, the negative clip is obtained from different timestamp $\mathbf{x}^{(i)}_{t'}$, where $t'\neq t$. This triplet of clips is passed through the utility model $f_T$ to achieve features denoted as $\mathbf{z}^{(i)}_{t}$, $\mathbf{\tilde{z}}^{(i)}_{t}$, and $\mathbf{z}^{(i)}_{t'}$. The proposed temporally-distinct triplet loss can be expressed as follows:
\begin{equation}\label{eq:triplet}
    \mathcal{L}^{(i)}_{D} = \max \{d(\mathbf{z}^{(i)}_{t}, \mathbf{\tilde{z}}^{(i)}_{t}) - d(\mathbf{z}^{(i)}_{t}, \mathbf{z}^{(i)}_{t'}) + {\mu}, 0\},
\end{equation}
where $d(u_j, v_j) = \left\lVert {\bf u}_j - {\bf v}_j \right\rVert_2$ is Euclidean distance between two vectors $\mathbf{u}$ and $\mathbf{v}$, and $\mu$ is the controllable margin hyperparameter to determine how far to push and pull features in the latent space.

We utilize this loss along with the standard cross-entropy action classification loss:

\begin{equation}\label{eq:crossentropy}
\mathcal{L}^{(i)}_{CE} = -\sum_{c=1}^{N_C} \mathbf{y}^{(i)}_{c}\log \mathbf{p}^{(i)}_{c},
\end{equation}

\noindent where $N_C$ is the total number of action classes of $\mathbb{D}_{action}$, $\mathbf{y}^{(i)}_{c}$ is the ground-truth label, and $\mathbf{p}^{(i)}_{c}$ is the prediction vector by utility model $f_T$.

Adding both temporal distinctiveness (Eq.~\ref{eq:triplet}) and action classification objective to our utility branch, our overall utility loss can be expressed as follows
\begin{equation}\label{eq:utility}
    \mathcal{L}_T = \mathcal{L}_{CE} + \omega*\mathcal{L}_{D},
\end{equation}
where $\omega$ hyperparameter is the weight of temporally-distinct triplet loss with respect to cross-entropy loss.

\noindent \textbf{Privacy (i.e. budget) Loss $\mathcal{L}_B$} We utilize the same self-supervised privacy loss from~\cite{spact}, which removes the private information by minimizing the agreement between the frames of the same video.

\noindent \textbf{Minimax Optimization} After reformulating the utility loss, we use the minimax optimization process similar to~\cite{spact}. It is a two-step iterative process that minimizes the utility loss and at the same time increases budget loss $\mathcal{L}_B$.  At the end of this optimization, we obtain the learned anonymization function ($f_A$) and utility video encoder ($f_T$).

\subsection{Privacy-preserving Anomaly Detection}
\label{sec:mgfn}
In order to detect anomalies within videos from $\mathbb{D}_{anomaly}$, we utilize the current state-of-the-art technique, Magnitude-Contrastive Glance-and-Focus Network (MGFN)~\cite{chen_mgfn_2022}. Similar to other recent works in anomaly detection, MGFN requires fixed features for each video from a pretrained video encoder for anomaly detection training. 

\noindent \textbf{Optimizing for Anomaly Detection} MGFN anomaly detection is comprised of 4 main losses: (1) a standard sigmoid cross-entropy loss $L_{sce}$ for snippet classification accuracy, (2) a temporal smoothing loss $L_{ts}$ \cite{sultani_real-world_2018} to encourage smoothness between feature representations of consecutive segments, (3) a sparsity term $L_{sp}$ \cite{sultani_real-world_2018} to discourage false positive anomalies, and (4) a novel magnitude contrastive loss $L_{mc}$ to learn scene-adaptive feature distributions across videos, all which help to train a model $f_{AD}$.

The training loss used in MGFN is compounded in the following equation: 
\begin{equation}
    L_{AD} = L_{sce} + \lambda_1L_{ts} + \lambda_2L_{sp} + \lambda_3L_{mc},
    \label{mgfn_loss}
\end{equation}
where $\lambda_1 = \lambda_2 = 1$, and $\lambda_3 = 0.001$. $f_{AD}$ outputs frame-level anomaly scores, which are used to calculate a final ROC AUC and AP for evaluation.

\noindent \textbf{Feature Extraction} In our privacy-preserving case, we cannot use $\mathbb{D}_{anomaly}$ directly for the feature extraction from the video encoder. We first anonymize each video ($X^i$) of the dataset through the learned anonymization function $f_A$ to get an anonymized dataset. For the feature extraction, we utilize the learned utility video encoder $f_T$. We denote this extracted set of anonymized features as $\mathbb{F}_{anomaly} = \{\,f_{T}(f_A(X^i))\; |\;  \forall X^i \in \mathbb{D}_{anomaly} \,\}$.

\subsection{Algorithm}
\label{sec:algo}
Let's consider the models $f_A$, $f_T$, $f_B$, $f_{AD}$, parameterized by $\theta_{A}$, $\theta_{T}$, $\theta_{B}$, and $f_{AD}$, respectively. ${\mathbb{D}_{action}}$ is the proxy action recognition dataset and ${\mathbb{D}_{anomaly}}$ is the downstream anomaly detection dataset. All training steps of our framework can be put together into a sophisticated form as algorithm~\ref{alg:algo1}.

\begin{algorithm}
\textbf{Inputs}:
            
        \hspace*{0.2cm} \textit{Datasets:} $\mathbb{D}_{action}$, $\mathbb{D}_{anomaly}$\\
        
        \hspace*{0.2cm} \textit{\#Epochs:} $max\_anon\_epoch$, $max\_ad\_epoch$\\
        
        \hspace*{0.2cm} \textit{Learning Rates: $\alpha_{AD}$, $\alpha_{B}$, $\alpha_T$}\\
        \hspace*{0.2cm} \textit{Hyperparameters: $\mu$, $\omega$}
     
\textbf{Output}: $\theta_{AD}$, $\theta_{A}$
\hrule
Model Initialization:

Initialize $\theta_T$ with Kinetics400 weights~\cite{carreira_quo_2018};

Initialize $\theta_B$ with SimCLR ImageNet weights~\cite{simclr}.

Initialize $\theta_A \gets \theta_A - \alpha_A \nabla_{\theta_A} (\mathcal{L}_{L1}(\theta_A))$(Ref. Eq.~\ref{eq:l1loss})

\hrule
Anonymization Training:

\For{$e_0 \gets 1$ \KwTo $max\_anon\_epoch$}
{   
    Step-1
    
    {\small$\theta_A \gets \theta_A - \alpha_A \nabla_{\theta_A} (\mathcal{L}_T(\theta_A,\theta_T) - 	\omega L_B(\theta_A, \theta_B))$}
    
    Step-2
    
    {\small$\theta_T \gets \theta_T - \alpha_T \nabla_{\theta_T} (\mathcal{L}_{T}(\theta_T, \theta_A)),$} (Ref. Eq.~\ref{eq:utility})
    {\small$\theta_B \gets \theta_B - \alpha_B \nabla_{\theta_B} (\mathcal{L}_{B}(\theta_B, \theta_A)).$}
}

\hrule

Feature Extraction on $\mathbb{D}_{anomaly}$:

$\mathbb{F}_{anomaly} = \{\,f_{T}(f_A(X^i))\; |\;  \forall X^i \in \mathbb{D}_{anomaly} \,\}$ 

\hrule

Privacy-Preserved Anomaly Detection Training:

\For{$e_0 \gets 1$ \KwTo $max\_ad\_epoch$}
{      
    {\small$\theta_{AD} \gets \theta_{AD} - \alpha_{AD} \nabla_{\theta_{AD}} (L_{AD}(\theta_{AD}, \mathbb{F}_{anomaly}))$}
}

\caption{TeD-SPAD Framework}
 \label{alg:algo1}
\end{algorithm}
\setlength{\textfloatsep}{12pt}

\section{Experiments}
\subsection{Datasets}

\noindent \textbf{UCF-Crime \cite{sultani_real-world_2018}} is the first large-scale weakly supervised video anomaly detection dataset. It contains 1,900 videos totaling 128 hours of untrimmed CCTV surveillance footage from a variety of different scenes. The videos contain 13 crime-based anomalies such as Arrest, Fighting, and Shoplifting in real world scenes. 

\noindent \textbf{XD-Violence \cite{wu2020not}} is currently the largest weakly supervised video anomaly detection dataset, totaling 217 hours of untrimmed video. All of its anomaly categories are related to violent activities. The videos are gathered from various types of cameras, movies, and games, resulting in a unique blend of scenes for increased difficulty.

\noindent \textbf{ShanghaiTech \cite{liu2018ano_pred}} is a medium-scale anomaly detection dataset containing videos covering 13 different scenes with various types of anomalies. While it was published as an unsupervised anomaly detection dataset, we use the weakly supervised rearrangement proposed by \cite{zhong_graph_2019}. 

\noindent \textbf{VISPR \cite{orekondy17iccv}} is an image dataset labelled with 68 privacy-related attributes, including gender, hair color, clothes, etc. It provides a multi-class classification problem for us to evaluate privacy-preservation on. The split we use for evaluation along with training details can found in \supp{Supp. Sec. B}.

\noindent \textbf{UCF101 \cite{soomro_ucf101_2012}} is a very common dataset in action recognition, and its relative simplicity makes it practical for learning an anonymization model on, demonstrated in \cite{spact}. In this work, split-1 of UCF101 is used as $\mathbb{D}_{action}$.

\begin{figure*}[h]
    \begin{subfigure}{0.32\textwidth}
        \includegraphics[clip, trim=0mm 0mm 0mm 2mm,width=\linewidth]{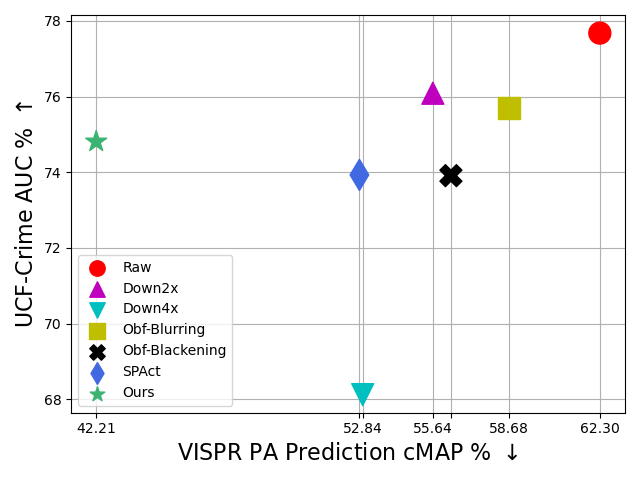}
        \caption{UCF-Crime}
    \end{subfigure}
    \begin{subfigure}{0.32\textwidth}
        \includegraphics[clip, trim=0mm 0mm 0mm 2mm,width=\linewidth]{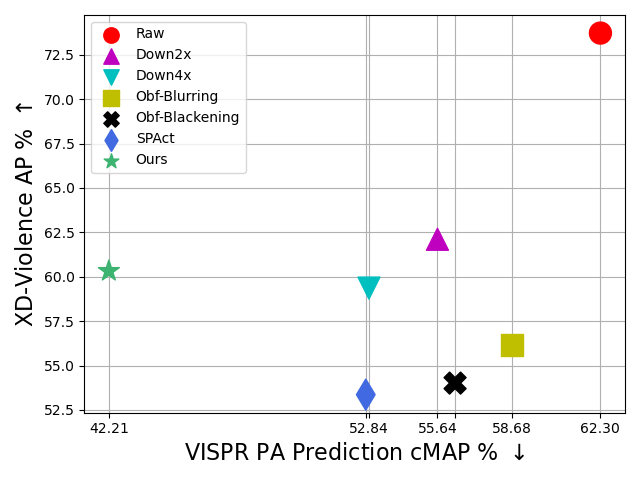}
        \caption{XD-Violence}
    \end{subfigure}
    \begin{subfigure}{0.32\textwidth}
        \includegraphics[clip, trim=0mm 0mm 0mm 2mm,width=\linewidth]{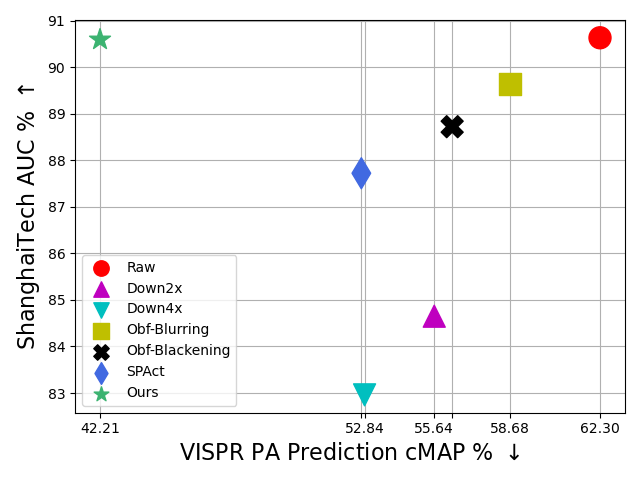}
        \caption{ShanghaiTech}
    \end{subfigure}
    \vspace{-2mm}

    \caption{Trade-off plots between anomaly detection benchmarks \cite{sultani_real-world_2018, wu2020not, liu2018ano_pred} AUC and VISPR \cite{orekondy17iccv} privacy attribute prediction cMAP for different privacy-preserving methods. Optimal trade-off point is top left of plot (higher AD performance, lower PA prediction ability).}
    \label{fig:privacy_tradeoff}
\end{figure*}

\subsection{Implementation Details}

\noindent \textbf{Network Architecture Details}
$f_A$ is a UNet \cite{ronneberger_u-net_2015} model that transforms raw input frames into anonymized frames. I3D \cite{carreira_quo_2018} is used for $f_T$, to first learn anonymized action classification, then to extract anonymized features. A ResNet-50 \cite{he_deep_2015} is our $f_B$ model during the anonymization training. $f_{AD}$ is MGFN \cite{chen_mgfn_2022}, which consists of a shortcut convolution, a self-attentional convolution, and a feed-forward network.

\noindent \textbf{Training Process Details} 
First we perform anonymization training for 80 epochs. Adam \cite{kingma2014adam} optimizer is used for all models with a learning rate is 1e-4 with a batch size of 8. Loss weight $\omega=0.1$ and margin $\mu=1$ in a default setting. We train the MGFN \cite{chen_mgfn_2022} model using default hyperparameters. $f_B$ evaluation uses a batch size of 32 and a base learning rate of 1e-3, which follows a linear warmup and a step-based scheduler that drops by a factor of 1/5 upon loss stagnation.
 
\noindent \textbf{Input Details} 
For all experiments, we crop each image to a scale of 0.8, then resize to input resolution $224\times224$. Clips consist of 16 frames and are sampled from a random start with a skip-rate of 2. For anonymization training we utilize standard augmentations like random erase, random crop, horizontal flip and random color jitter. To maintain temporal consistency, augmentations are applied equally to every frame within each clip.

\noindent \textbf{Feature Extraction}
Given a raw input video $\mathbf{X^i} \in \mathbb{D}_{anomaly}$, we first extract $S$ clips, where $S$ is the amount of non-overlapping 16-frame clips in $\mathbf{X}$. We sequentially pass each clip $S^{(j)}$ first through our anonymizer $f_A$, then our feature extractor $f_T$. Extracted features $f_T(f_A(\mathbf{X}^i))$ have the shape $S$ x $C$, where $C$ is the dimensionality of the feature vector. Specifically, features are extracted following the average pooling after the \textit{mix\_5c} layer of I3D and have a dimensionality of 2048.

\noindent \textbf{Evaluation Protocol and Performance Metrics}
To evaluate the learned anonymization function $f_A$, we follow standard protocols of cross-dataset evaluation~\cite{spact, wang2019privacy}. In this protocol, testing videos of $\mathbb{D}_{anomaly}$ are anonymized by $f_A$, and frame-level predictions are obtained through the $f_T$ and $f_{AD}$. The calculated ROC AUC is used to evaluate performance on UCF-Crime and ShanghaiTech, and AP is used for XD-Violence. A higher AUC/AP is considered a more accurate anomaly localization.  In order to evaluate the privacy leakage, the learned $f_A$ is utilized to anonymize the privacy dataset $\mathbb{D}_{privacy}$ to train and evaluate a target privacy model $f'_B$. Privacy leakage is measured in terms of the performance of the target $f'_{B}$ on the test set of $\mathbb{D}_{privacy}$. Since the privacy dataset is multi-label, the privacy leakage is measured in terms of mean average precision averaged across classes (cMAP). 

\noindent More implementation details are in \supp{Supp. Sec. C.}

\subsection{Privacy-Preserving Baselines}
We run well-known self-supervised privacy-preservation techniques for video anomaly detection. In order to maintain a fair comparison across methods, we utilize the exact same network architectures and training process.

\noindent \textbf{Downsampling Baselines} 
For \texttt{Downsample-2x} and \texttt{Downsample-4x}, we simply resize the input frames to a lower resolution by a factor of 2 ($112\times112$) and 4 ($56\times56$). 

\noindent \textbf{Object-Detector Based Obfuscation Baselines}
Obfuscation techniques are based on first detecting the person, followed by removing (i.e blackening) or blurring them. Both obfuscation techniques use MS-COCO \cite{mscoco} pretrained YOLO \cite{yolo} object detector to obtain bounding boxes for \texttt{person} object class. We utilize YOLOv5\footnote{\href{https://github.com/ultralytics/yolov5}{https://github.com/ultralytics/yolov5}} implementation with yolov5x as backbone. The detected bounding boxes are assigned to pixel value 0 for the \texttt{Blackening}-based baseline. For the \texttt{Blurring}-based baselines, a Gaussian filter with kernel $k=13$ and variance $\sigma=10$ is utilized.

\noindent \textbf{SPAct~\cite{spact} Baseline} We utilize official implementation\footnote{\href{https://github.com/DAVEISHAN/SPAct}{https://github.com/DAVEISHAN/SPAct}}. For a fair comparison with our method, we utilize the exact same utility model I3D and privacy model ResNet-50.

\subsection{Evaluation on Benchmark Anomaly Datasets}

We compare prior privacy-preserving methods to our method on 3 well-known anomaly detection benchmark datasets. Since privacy-preservation deals with both utility (i.e. anomaly) and privacy, we show results in form of a trade-off plot as presented in Fig.~\ref{fig:privacy_tradeoff}. Compared to the prior best method~\cite{spact}, our method is able to remove 19.9\% more privacy with a slightly better utility score (1.19\%). This strongly supports our claim that promoting temporal distinctiveness during anonymization better aligns with anomaly detection tasks. Numeric data behind Fig.~\ref{fig:privacy_tradeoff} plots can be found in \supp{Supp. Sec. D}.

\begin{figure*}
\centering
    \begin{subfigure}{0.445\textwidth}
    \centering
        \includegraphics[width=\textwidth]{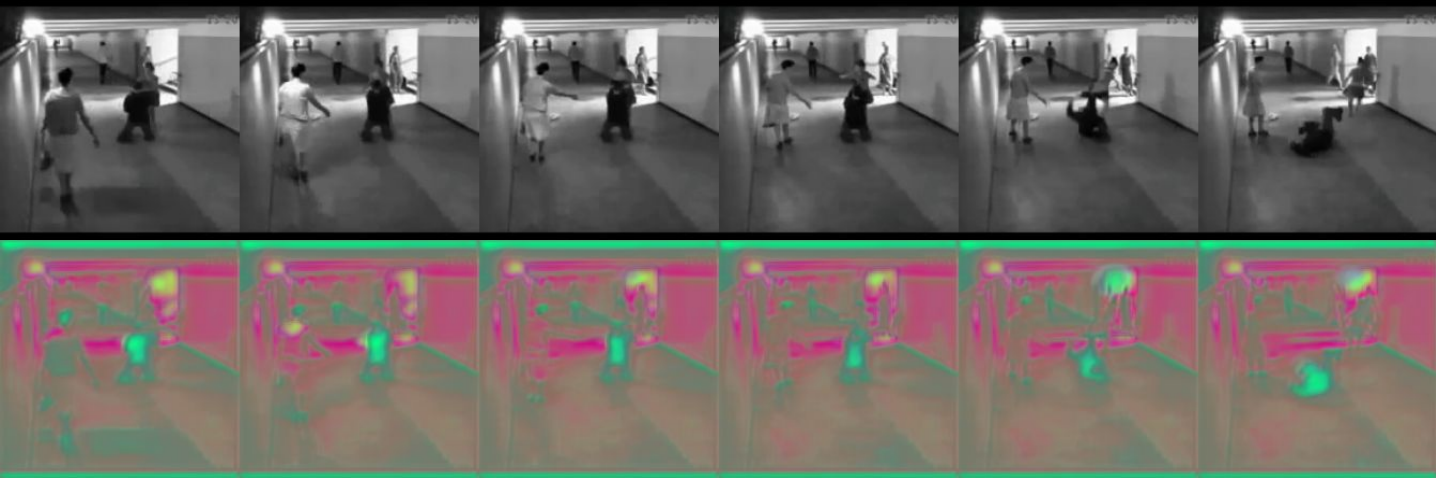}
        \caption{Fighting018\_x264.mp4}
        \label{qual_a}
    \end{subfigure}
    \hfill
    \begin{subfigure}{0.445\textwidth}
        \centering
        \includegraphics[width=\textwidth]{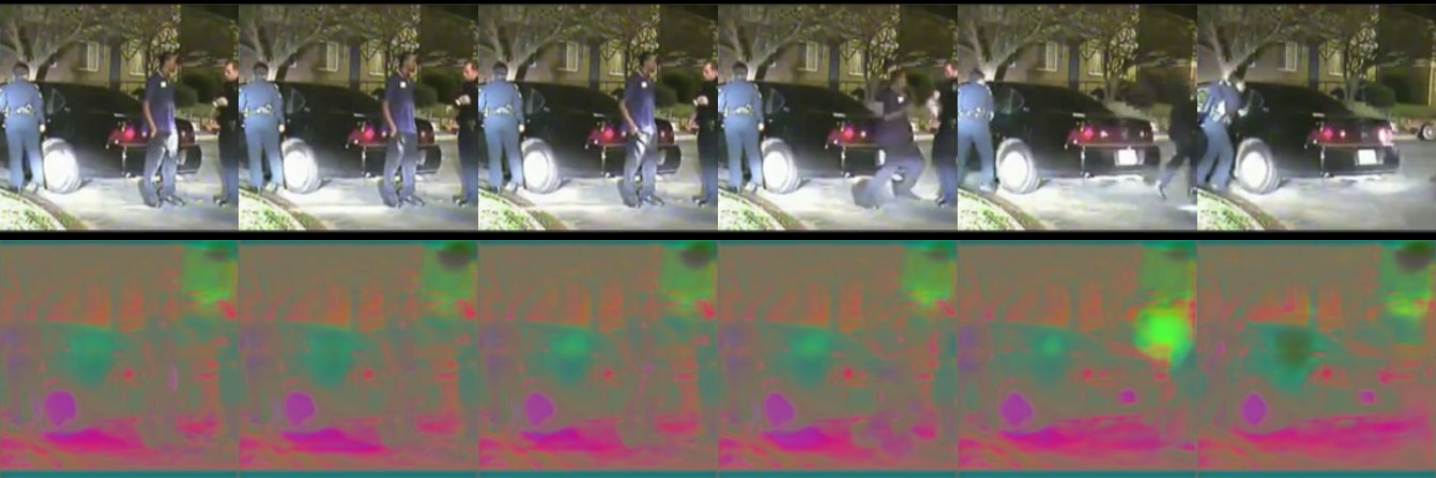}
        \caption{Shooting040\_x264.mp4}
        \label{qual_b}
    \end{subfigure}
    \hfill
    \begin{subfigure}{0.445\textwidth}
        \centering
        \includegraphics[width=\columnwidth]{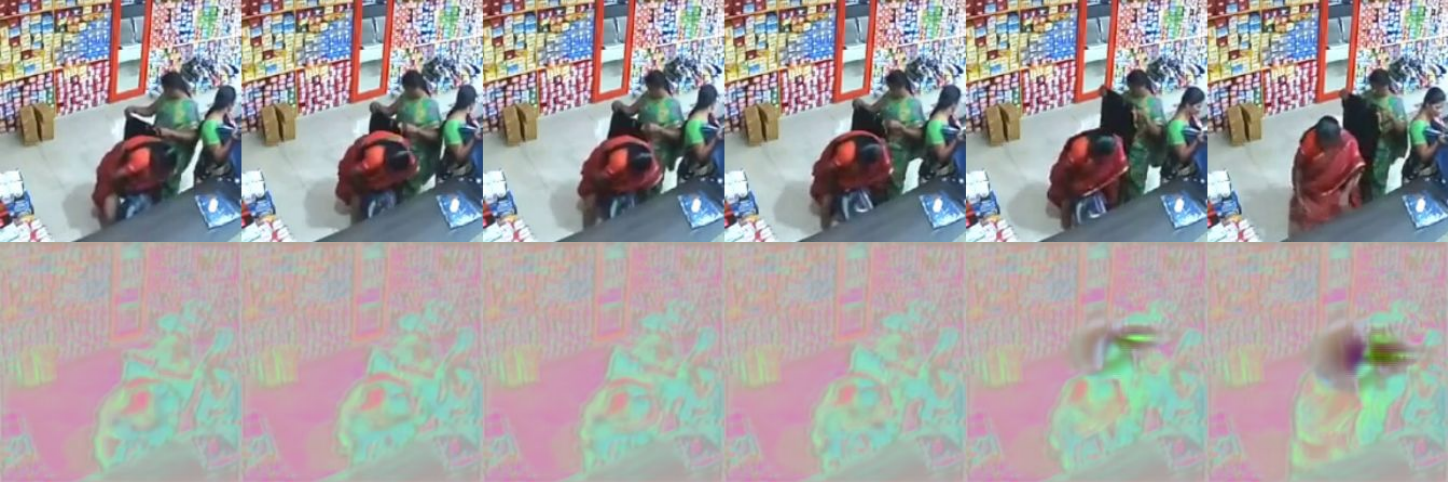}
        \caption{Shoplifting048\_x264.mp4}
        \label{qual_c}
    \end{subfigure}
    \hfill
    \begin{subfigure}{0.445\textwidth}
        \centering
        \includegraphics[width=\columnwidth]{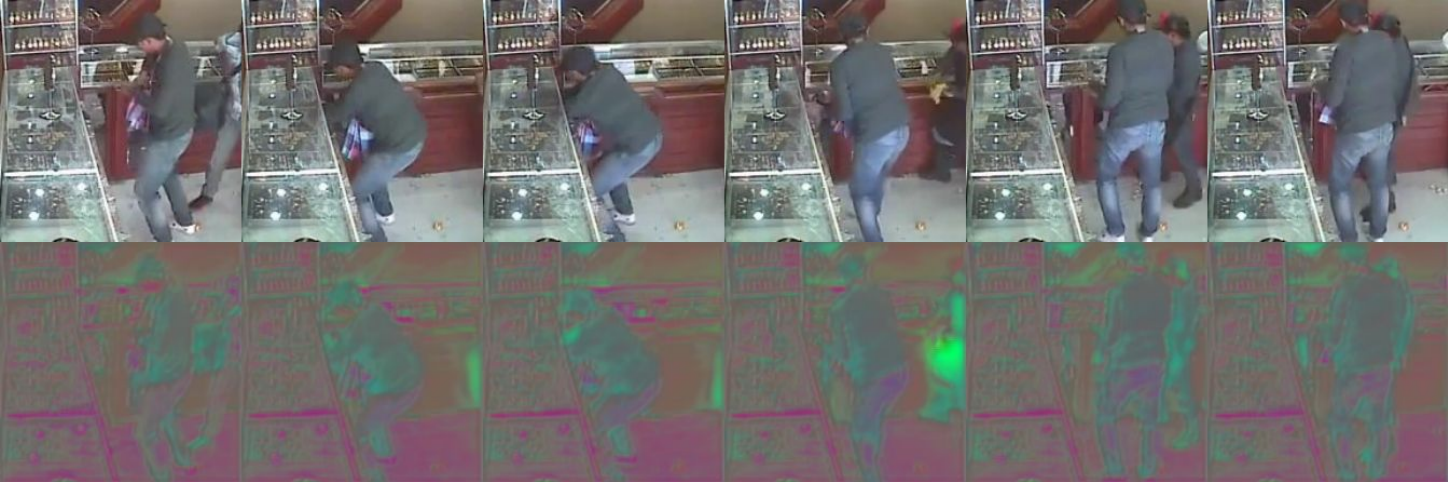}
        \caption{Robbery023\_x264.mp4}
        \label{qual_d}
    \end{subfigure}

    \vspace{-2mm}
    \caption{Qualitative results of our anonymization training on UCF-Crime \cite{sultani_real-world_2018} dataset. For each case, the top row shows the raw video frames, and the bottom row shows the frames after being passed through the anonymizer.}    
    \label{fig:qualviz}

\end{figure*}

\begin{figure*}
    \centering
    \includegraphics[width=0.70\textwidth]{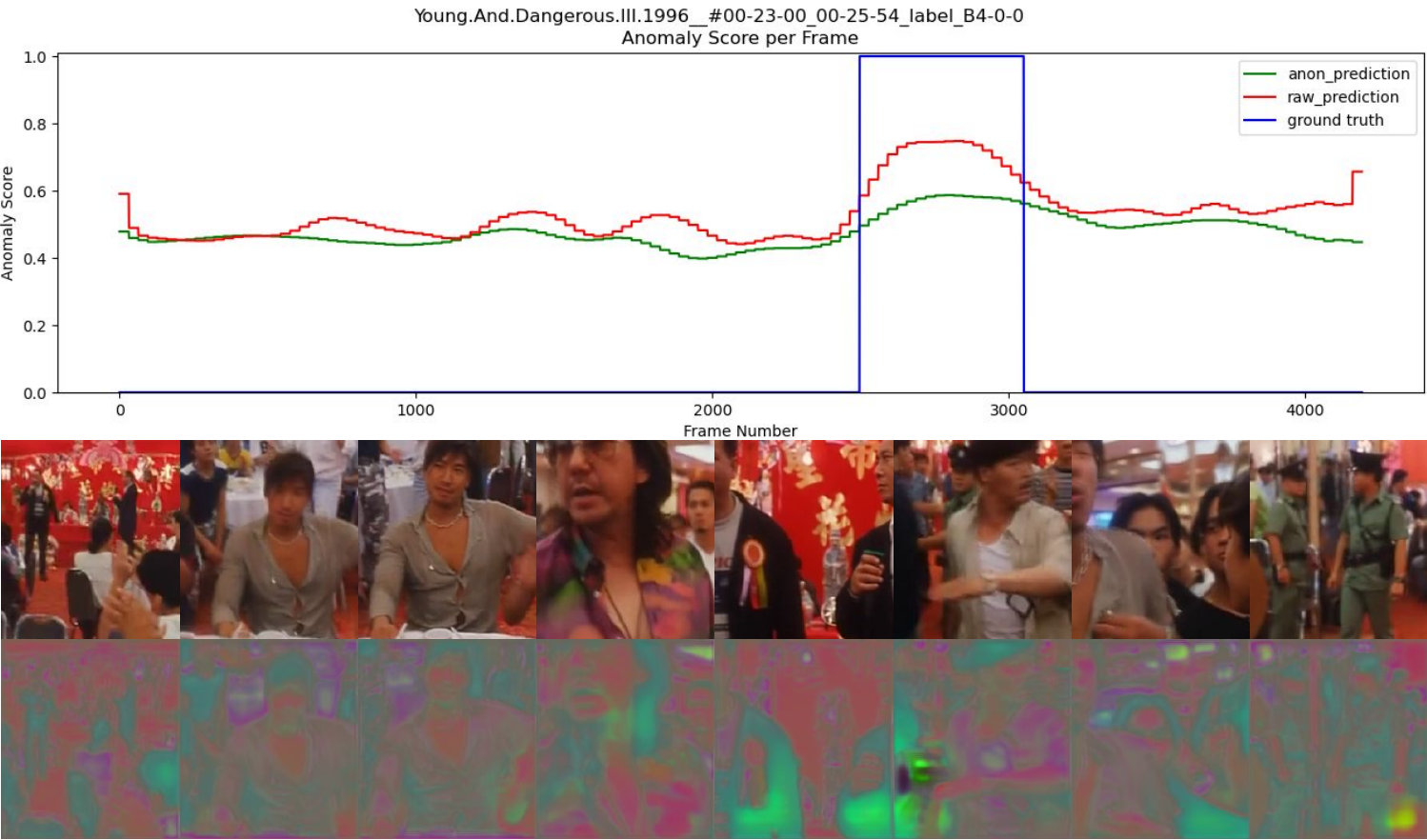}
    \caption{Frame-level anomaly score plot from XD-Violence. \textcolor{ForestGreen}{Green} line shows our anonymized model, \textcolor{red}{red} line is the raw input model, both compared to the \textcolor{blue}{blue} ground truth line. The below visualizations shows uniformly sampled frames from the video.}
    \label{fig:yad_plot}
\end{figure*}

\subsection{Qualitative Results}

Figure \ref{fig:qualviz} shows visual examples of the model outputs in different videos. We note that to the human eye, it is difficult to tell what is going on in each video, yet the anomaly detection model is still able to achieve high performance. Fig. \ref{qual_a}-\ref{qual_d} include human subjects, yet none of their private attributes such as face and clothing are visible, and therefore cannot be used to make unfair decisions. In Fig. \ref{qual_c}, \ref{qual_d}, the background shop can still be seen, which is useful context for identifying the shoplifting and robbery anomalies. 

Figure~\ref{fig:yad_plot} is a graph of the anomaly score output from both the raw and anonymized video models for each frame of an input video. The blue box shows the ground truth anomalous frames. We see that the green anonymized anomaly scores are similar to the red raw scores while still increasing the anomaly scores on the ground truth anomaly, showing that our anonymization technique maintains utility performance. See \supp{Supp. Sec. D} for more qualitative results.

\subsection{Evidence for Privacy Leakage at Feature-level}
In anomaly feature representation learning, the anomaly detection algorithms do not directly work with the input videos, the videos are first passed through an action classifier to compute features. Even though the action recognition model sees the original videos, it is not certain whether the private information gets passed to the features. In order to confirm this, we create a simple fully connected network to predict VISPR private attributes, in the same fashion that we evaluate privacy for our other experiments. We stack the same VISPR image 16 times to create a video clip, then extract the clip features through $f_T$. Our baseline uses unmodified input images passed through a pretrained Kinetics400 \cite{kay_kinetics_2017} I3D \cite{carreira_quo_2018} model, with the other experiments using a paired anonymizer and tuned I3D model. Detailed explanations of this process are found in \supp{Supp. Sec. C}. Through experimentation, we find that {\em the action classifier latent features do in fact leak private information, therefore this private information gets passed into the anomaly detector}. Table~\ref{tab:feature_privacy} demonstrates empirical evidence of this.

\begin{table}
\begin{center}
\resizebox{\columnwidth}{!}{\begin{tabular}{l c c}
\hline

\hline

\hline\\[-3mm]
& \bf VISPR & \\
Feature Extraction Model & Privacy cMAP (\%) & Privacy Reduction (\%) \\
\hline
\textcolor{red}{Kinetics400 pretrained} & \textcolor{red}{63.15} & \textcolor{red}{0.0} \\
SPAct \cite{spact} Anonymized & 55.60 & 11.96\\
Our Anonymized & 52.30 & 17.18 \\
\hline

\hline

\hline\\[-3mm]
\end{tabular}}
\end{center}
\vspace{-4mm}

\caption{Quantitative evidence of action classification model features leaking privacy. Features from each model used to predict privacy attributes. Red indicates higher privacy leakage.}
\label{tab:feature_privacy}
\end{table}

\label{feature_privacy_leak}

\subsection{Ablation Study}

\noindent \textbf{Effect of different utility losses $\mathcal{L}_{T}$} We study the effect of different utility losses during the anonymization process on the final privacy vs. utility anomaly detection performance in Table~\ref{tab:utilitylosses}. From \texttt{row-(a,b)}, we see a clear reduction in privacy at no cost of anomaly performance; which demonstrates the effectiveness of our proposed temporally-distinct triplet loss $\mathcal{L}_{D}$. 

\begin{table}[b]
\centering
\begingroup
\setlength{\tabcolsep}{2pt}
\begin{tabular}{llcc} 
\hline

\hline

\hline\\[-3mm]
\multirow{3}{*}{} & \multirow{3}{*}{\begin{tabular}[c]{@{}l@{}}Utility Loss during \\Anonymization ($\mathcal{L}_T$)\end{tabular}} & \textbf{VISPR} & \textbf{UCF-Crime}  \\
                  &                                                                                                   & Privacy        & Anomaly              \\
                  &                                                                                                   & cMAP(\%)($\downarrow$)           & AUC(\%)($\uparrow$)                  \\ 
\hline
\texttt{(a)}                 & $\mathcal{L}_{CE}$                                                                                           & 52.71          & 73.93                \\
\texttt{(b)}                 & $\mathcal{L}_{CE} + \mathcal{L}_{D}$                                                                                          & \textbf{42.21}          & \textbf{74.81}                \\
\texttt{(c)}                 & $\mathcal{L}_{CE} + \mathcal{L}_{I}$                                                                                        & 45.64          & 69.52                \\
\hline

\hline

\hline\\[-4mm]
\end{tabular}
\endgroup
\caption{Ablation with different utility losses during the anonymization process. Bold indicates best trade-off.}
\label{tab:utilitylosses}
\end{table}

To this extent, we also implement the contrary objective to our $\mathcal{L}_{D}$ which promotes temporal-invariance $\mathcal{L}_{I}$ in the utility branch of the anonymization training. We implement this using well-known self-supervised works~\cite{cvrl,dave_tclr_2022}. From \texttt{row-(c)} we see that temporal-invariance objective is not well-suited for anomaly detection utility tasks and results in a significant drop of \textbf{6\%}. We provide extensive experiments with $\mathcal{L}_{I}$ and its explanation in \supp{Supp. Sec. D}.

\noindent \textbf{Effect of different temporal distinctiveness objectives:}

Our objective of temporal distinctiveness can also be achieved through a contrastive loss. To this end we utilize the implementation of local-local temporal contrastive loss (LLTC) of~\cite{dave_tclr_2022}. It achieves 75.06\% AUC at 42.44\% cMAP (Table~\ref{tab:ablation_distinctiveness}), which is closely comparable to our results using triplet loss. It is important to note that LLTC loss significantly increases the computation (GPU memory requirement) compared to triplet loss. It requires 8 clips (4 clips $\times$ 2 augmented view) which results in 132.45G FLOPs compared to 3 clips of triplet loss requiring only 49.67G FLOPs in our experimental setup. 

\begin{table}[h]
\begin{center}
\begin{tabular}{l c c}
\hline

\hline

\hline\\[-3mm]
Temporal & \bf VISPR & \bf UCF-Crime \\
Distinctiveness & Privacy & Anomaly \\
Loss $\mathcal{L}_D$ & cMAP(\%)($\downarrow$) & AUC(\%)($\uparrow$) \\
\hline
\rowcolor[rgb]{0.922,0.922,0.922} None & 52.71 & 73.93 \\
LLTC \cite{dave_tclr_2022} & 42.44 & \bf 75.06 \\
Triplet & \bf 42.21 & 74.81 \\
\hline

\hline

\hline\\[-3mm]
\end{tabular}
\end{center}
\vspace{-3mm}
\caption{Comparison of different loss formulations to encourage temporal distinctiveness during the anonymization process. Both achieved similar results, with the triplet loss requiring significantly less computation.}

\label{tab:ablation_distinctiveness}
\end{table}

\noindent \textbf{Relative Weightage of $\mathcal{L}_{D}$ } Here we test the effect of changing the weight of the additional temporally-distinct triplet loss. Table~\ref{tab:ablation_lossweight} shows that weighting the loss at 0.1 achieves our best results of 32.25\% relative increase in privacy with only a 3.69\% reduction in utility performance. Without the enforced temporal distinctiveness, the utility model is limited by the quality of the reconstructed anonymized videos. The improper weighting of the temporal loss interferes with the classifier ability of the model, which can also harm the anonymization process. This suggests that action recognition loss $\mathcal{L}_{CE}$ is still an important utility task for anomaly detection performance.

\begin{table}
\begin{center}
\begin{tabular}{l c c}
\hline

\hline

\hline\\[-3mm]
Triplet & \bf VISPR & \bf UCF-Crime \\
Temporal & Privacy & Anomaly \\
Loss Weight $\omega$ & cMAP(\%)($\downarrow$) & AUC(\%)($\uparrow$) \\
\hline
\rowcolor[rgb]{0.922,0.922,0.922} 0 & 52.71 & 73.93 \\
0.01 & 53.84 & 72.53 \\
\bf 0.1 & \bf 42.21 & \bf 74.81 \\
1.0 & 55.26 & 70.56 \\
10.0 & 51.67 & 69.27 \\
\hline

\hline

\hline\\[-3mm]
\end{tabular}
\end{center}
\vspace{-4mm}
\caption{Comparison of using different loss weights of the temporally-distinct triplet loss during the anonymization process. The margin hyperparameter of the temporal triplet loss for each experiment was 1. Bold indicates best trade-off.}
\label{tab:ablation_lossweight}
\end{table}

\noindent \textbf{Effect of the margin in $\mathcal{L}_{D}$} The proposed $\mathcal{L}_D$ temporally-distinct triplet loss uses a margin hyperparameter $\mu$ to allow for adjustments to the contrastive distance in the latent feature space. The only requirement is that $\mu > 0$. The intuition here is that a larger margin enforces greater feature spacing. As Table~\ref{tab:ablation_margin} shows, we empirically found that setting $\mu = 1$ gives us the most robust results. A lower margin results in less temporally-distinct representations, which makes distinguishing between normal and anomalous features more difficult. On the other hand, increasing the $\mu=2$ results in more difficult temporal triplet loss (i.e a very high temporal distinctiveness) which may not align well with anomaly detection task. 

\begin{table}[h]
\begin{center}

\begin{tabular}{l c c}
\hline

\hline

\hline\\[-3mm]
Triplet & \bf VISPR & \bf UCF-Crime \\
Temporal & Privacy & Anomaly \\
Loss Margin $\mu$ & cMAP(\%)($\downarrow$) & AUC(\%)($\uparrow$) \\
\hline
0.5 & 49.78 & 62.12 \\
\bf 1.0 & \bf 42.21 & \bf 74.81 \\
2.0 & 58.86 & 67.30 \\
\hline

\hline

\hline\\[-3mm]
\end{tabular}
\end{center}
\vspace{-6mm}

\caption{Comparison of using different hyperparameter margins of the temporally-distinct triplet loss during the anonymization process. Bold indicates default setup.}
\label{tab:ablation_margin}
\end{table}

\noindent \textbf{Effect of temporal sampling in $\mathcal{L}_{D}$} The proposed triplet loss (Eq.~\ref{eq:triplet}) forms the negative from the clip $\mathbf{x}^{(i)}_{t'}$  of a different timestamp $t'$. Distance between the timestamp of negative and anchor clip $t-t'$ is an important aspect to define temporal distinctiveness. We perform experiments with various distances as shown in Table~\ref{tab:ablation_distance}. In our default setting, we use random distance as shown in the first row. From the second row, we can say that a smaller distance leads to a better anomaly score with a slight degradation in protecting privacy. At the same time, the third row suggests that enforcing temporal distinctiveness at higher distances leads to better privacy protection but at the cost of anomaly performance. This distance hyperparameter may be used as a tuning parameter to get different operating points of privacy vs. anomaly trade-off. 

\begin{table}[h]
\begin{center}
\begin{tabular}{l c c}
\hline

\hline

\hline\\[-3mm]
Negative & \bf VISPR & \bf UCF-Crime \\
Clip & Privacy & Anomaly \\
Distance & cMAP(\%)($\downarrow$) & AUC(\%)($\uparrow$) \\
\hline
\textbf{Random} & \textbf{42.21} & \textbf{74.81} \\
8 & 46.34 & 76.12 \\
32 & 28.69 & 70.97 \\
\hline

\hline

\hline\\[-3mm]
\end{tabular}
\end{center}
\vspace{-3mm}

\caption{Comparison of enforcing set clip sampling distance during the anonymization process. The margin hyperparameter of the temporal triplet loss for each experiment was 1.}
\label{tab:ablation_distance}

\end{table}

\section{Conclusion}
In this paper, we highlight the importance of privacy, a previously neglected aspect of video anomaly detection. We present TeD-SPAD, a framework for applying Temporal Distinctiveness to Self-supervised Privacy-preserving video Anomaly Detection. TeD-SPAD demonstrates the effectiveness of using a temporally-distinct triplet loss while anonymizing an action recognition model, as it enhances feature representation temporal distinctiveness, which complements the downstream anomaly detection model. By effectively destroying spatial private information, we remove the model's ability to use this information in its decision-making process. As a future research direction, this framework can be extended to other tasks, such as spatio-temporal anomaly detection. The anonymizing encoder-decoder may also be made more powerful with techniques using recent masked image modeling. It is our hope that this work contributes to the development of more responsible and unbiased automated anomaly detection systems.

\section{Acknowledgements}

This work was supported in part by the National Science Foundation (NSF) and Center for Smart Streetscapes (CS3) under NSF Cooperative Agreement No. EEC-2133516.

{\small
\bibliographystyle{ieee_fullname}
\bibliography{egbib}
}

\newpage
\mbox{~}
\clearpage

\appendix
\section{Supplementary Overview}

Section~\ref{sec:datasets}: Dataset details

Section~\ref{sec:imp_details}: Implementation details

Section~\ref{sec:results}: Additional results

\section{Dataset Details}
\label{sec:datasets}

\noindent \textbf{UCF-Crime \cite{sultani_real-world_2018}} contains 1,900 (950 normal, 950 anomalous) videos with 13 different crime-based anomalies, for a total of 128 hours. The labels are included at the video level, indicating whether or not the video contains at least one anomalous event. The footage comes from real-life CCTV surveillance cameras in a variety of scenes. The average video contains 7,247 frames, which is $\approx$3 minutes at 30fps. The training set has a total of 800 normal videos and 810 anomalous videos, and the testing set has 150 normal and 140 anomalous videos. Both sets contain examples of all anomaly categories, with some videos having multiple anomalies.\\

\noindent \textbf{XD-Violence \cite{wu2020not}} contains 4,754 (2405 normal, 2349 anomalous) videos with 6 different anomaly categories, total 217 hours of untrimmed footage, making it the largest weakly supervised video anomaly detection dataset. The labels are also at the video level, except they allow for each video to have more than one anomaly label. The videos also contain audio signals to allow for multi-modal anomaly detection. The videos are gathered from various types of cameras, movies, and games, resulting in a unique blend of scenes for increased difficulty. The training set contains 3,954 videos while the test set has 800 videos total, 500 anomalous and 300 normal.\\

\noindent \textbf{ShanghaiTech \cite{liu2018ano_pred}} contains 437 videos in 13 different scenes with a total of 130 anomalous events. The training set includes 330 videos while the test set includes 107. Out of a total of $\approx$317,400 frames in the dataset, 17,900 are anomalous. Each anomaly also contains a pixel-level location for anomaly localization. It was published as an unsupervised anomaly detection dataset, but Zhong~\etal \cite{zhong_graph_2019} proposed a weakly supervised rearrangement, which is used in this work.\\

\noindent \textbf{VISPR \cite{orekondy17iccv}} is a visual privacy image dataset containing ~22k public Flickr images labelled with 68 different private attributes. Private attributes are determined by personally identifiable information as considered in the US Privacy Act of 1974 and the EU Data Protection Directive 95/46/EC \cite{noauthor_directive_1995}. The training and testings sets contain 10,000 and 8,000 images, respectively. For ease of comparison, we use the same VISPR attribute split used in \cite{wang2019privacy, spact}, seen in Table~\ref{tab:vispr_split}.\\

\begin{table}
\centering
\begin{tblr}{
  cells = {c},
  cell{1}{1} = {c=2}{},
  hline{1-2} = {-}{},
  hline{3} = {-}{},
  hline{10} = {-}{},
}
\textbf{VISPR1 \cite{wang2019privacy, spact}}   &                             \\
\bf Label              & \bf Description             \\
\tt a17\_color         & skin color                  \\
\tt a4\_gender         & gender                      \\
\tt a9\_face\_complete & full face visible           \\
\tt a10\_face\_partial & part of face visible        \\
\tt a12\_semi\_nudity  & partial nudity              \\
\tt a64\_rel\_personal & shows personal relationship \\
\tt a65\_rel\_soci     & shows social relationship   \\
\end{tblr}
\caption{Privacy attributes from subset of VISPR \cite{orekondy17iccv} labels as used in previous works.}
\label{tab:vispr_split}
\end{table}

\noindent \textbf{UCF101 \cite{soomro_ucf101_2012}} contains 13,320 videos in 101 different human action categories. In the default setting, split-1 is used. Each video shows the action directly with no filler, so the average video length is 7.21s. \\

\noindent \textbf{Kinetics400 \cite{kay_kinetics_2017}} is used as the standard video dataset for action classifier pretraining. The dataset contains a total of 306,245 videos, with over 400 examples of each of the 400 human action classes.

\begin{table*}
\begin{center}
{\begin{tabular}{l c c c c}
\hline

\hline

\hline\\[-3mm]
& \bf VISPR & \bf UCF-Crime & \bf XD-Violence & \bf ShanghaiTech \\
Method & Privacy & Anomaly & Anomaly & Anomaly \\
& cMAP(\%)($\downarrow$) & AUC(\%)($\uparrow$) & AP(\%)($\uparrow$) & AUC(\%)($\uparrow$)\\
\hline
Raw data & 62.30 & 77.68 & 73.72 & 90.63 \\
Downsample-2x & 55.64 \textcolor{ForestGreen}{$\downarrow$10.69\%} & 76.09 \textcolor{red}{$\downarrow$2.05\%} & 62.11 \textcolor{red}{$\downarrow$15.75\%} & 84.65 \textcolor{red}{$\downarrow$6.60\%} \\
Downsample-4x & 52.84 \textcolor{ForestGreen}{$\downarrow$15.18\%} & 68.12 \textcolor{red}{$\downarrow$12.31\%} & 59.36 \textcolor{red}{$\downarrow$19.48\%} & 82.96 \textcolor{red}{$\downarrow$8.46\%} \\
Obf-Blurring & 58.68 \textcolor{ForestGreen}{$\downarrow$5.81\%} & 75.69 \textcolor{red}{$\downarrow$2.56\%} & 59.36 \textcolor{red}{$\downarrow$23.81\%} & 89.63 \textcolor{red}{$\downarrow$1.10\%} \\
Obf-Blackening & 56.36 \textcolor{ForestGreen}{$\downarrow$9.53\%} & 73.91 \textcolor{red}{$\downarrow$4.85\%} & 56.17 \textcolor{red}{$\downarrow$26.74\%} & 88.72 \textcolor{red}{$\downarrow$2.11\%} \\
SPAct \cite{spact} & 52.71 \textcolor{ForestGreen}{$\downarrow$15.39\%} & 73.93 \textcolor{red}{$\downarrow$4.83\%} & 53.36 \textcolor{red}{$\downarrow$27.62\%} & 87.72 \textcolor{red}{$\downarrow$3.21\%} \\
\bf Ours & \bf 42.21 \textcolor{ForestGreen}{$\downarrow$32.25\%} & \bf 74.81 \textcolor{red}{$\downarrow$3.69\%} & \bf 60.32 \textcolor{red}{$\downarrow$18.18\%} & \bf 90.59 \textcolor{red}{$\downarrow$0.04\%} \\
\hline

\hline

\hline\\[-3mm]
\end{tabular}}
\end{center}
\vspace{-3mm}

\caption{Comparison with different privacy-preservation methods on UCF-Crime, XD-Violence and ShanghaiTech anomaly detection. Bold indicates the best trade-off results. Trade-off plots are shown in \main{main paper Fig. 3}.  Downward arrows {\textcolor{red}{$\downarrow$}} and {\textcolor{ForestGreen}{$\downarrow$}} show the relative percent change compared to the raw data.}
\label{tab:main_table}
\end{table*}

\section{Implementation Details}
\label{sec:imp_details}

All code is implemented using the PyTorch~\cite{paszke2019pytorch} library. 

\subsection{Feature-level Privacy Leakage Tester}
To test privacy leakage at the feature-level \main{(main paper Sec. 4.6)}, we create a simple fully connected model $f_P$ consisting of 5 layers: Linear(2048, 2048) $\rightarrow$ Linear(2048, 1028) $\rightarrow$ Linear(1028, 1028) $\rightarrow$ Linear(1028, 512) $\rightarrow$ Linear(512, 7). This model is trained for 50 epochs with a cross-entropy with logits loss and Adam \cite{kingma2014adam} optimizer at a learning rate of 1e-4. Images are augmented similar to the test set images, then stacked 16 times to resemble a video for feature extraction input. The set of 2048 dimensional features $\mathbb{F}_{anomaly}$ from the I3D $f_T$ model is directly input to this privacy leakage training model. 

\subsection{Anonymization Process}

\subsubsection{Input Augmentations}
We utilize standard augmentations following~\cite{spact}. During training, we utilize random cropping, scaling, color jittering, erasing, and horizontal flipping. During inference, we utilize center crop with a scale of 0.8.

\subsection{MGFN}

We use the official MGFN \cite{chen_mgfn_2022} implementation\footnote{\href{https://github.com/carolchenyx/MGFN}{https://github.com/carolchenyx/MGFN}} for anomaly detection evaluation. Besides using only single crop features instead of ten-crop, we use their exact hyperparameters. The residual feature norm for each segment is appended with a weight of 0.1. To help mitigate potential noise, the top-k clips are considered in the loss instead of top-1, with $k=3$. The feature dropout rate in training is 0.7. The optimizer employed is Adam \cite{kingma2014adam}, starting with a learning rate of 0.001 with a weight decay of 0.0005, trained for up to 1000 epochs with a batch size of 16.

For reference, the compound MGFN loss function is: 
\begin{equation}
    L_{AD} = L_{sce} + \lambda_1L_{ts} + \lambda_2L_{sp} + \lambda_3L_{mc},
    \label{mgfn_loss}
\end{equation}
where $\lambda_1 = \lambda_2 = 1$, and $\lambda_3 = 0.001$.

The base loss starts with standard sigmoid cross entropy loss:
\begin{equation}
    L_{sce} = -ylog(s^{i,j})-(1-y)log(1-s^{i,j}),
    \label{sig_ce}
\end{equation}
where y is video-level label ($y=1$ is anomaly, $y=0$ is normal), $s^{i,j}$ is the computed anomaly score for frames $i$ in segment $j$.

Sultani et al. \cite{sultani_real-world_2018} proposed the use of a temporal smoothness $L_{ts} = \sum^{(n-1)}_i(f(V^i_a) - f(V^{i+1}_a))^2$ and a sparsity term $L_{sp} = \sum^n_if(V^i_a),$ where $f(V^i_a)$ is the extracted features for segment $i$ of anomalous video $V_a$. These encourage infrequent anomaly detections and smoothness between representations of sequential video segments.

MGFN also includes a feature amplification mechanism paired with a magnitude contrastive (MC) loss (Eq. \ref{mag_con}) to better enhance feature separability both within videos and between videos.
The MC loss is formulated as follows:
\begin{equation}
    \begin{split}
        L_{mc} = \sum^{B/2}_{p,q=0}(1-l)(D(M^p_n, M^q_n)) + \sum^B_{u,v=B/2}(1-l)(D \\
        (M^u_a,M^v_a)) + \sum^{B/2}_{p=0}\sum^B_{u=B/2}l(Margin-D(M^p_n,M^u_a)),
    \end{split}
    \label{mag_con}
\end{equation}
where $B$ is the batch size, $M$ is the feature magnitude of the corresponding segment, $D(\cdot,\cdot)$ is a distance function, and $l$ is an indicator function. For more details about this loss, refer to \cite{chen_mgfn_2022}.

\begin{figure*}
\centering
    \includegraphics[width=0.85\textwidth]{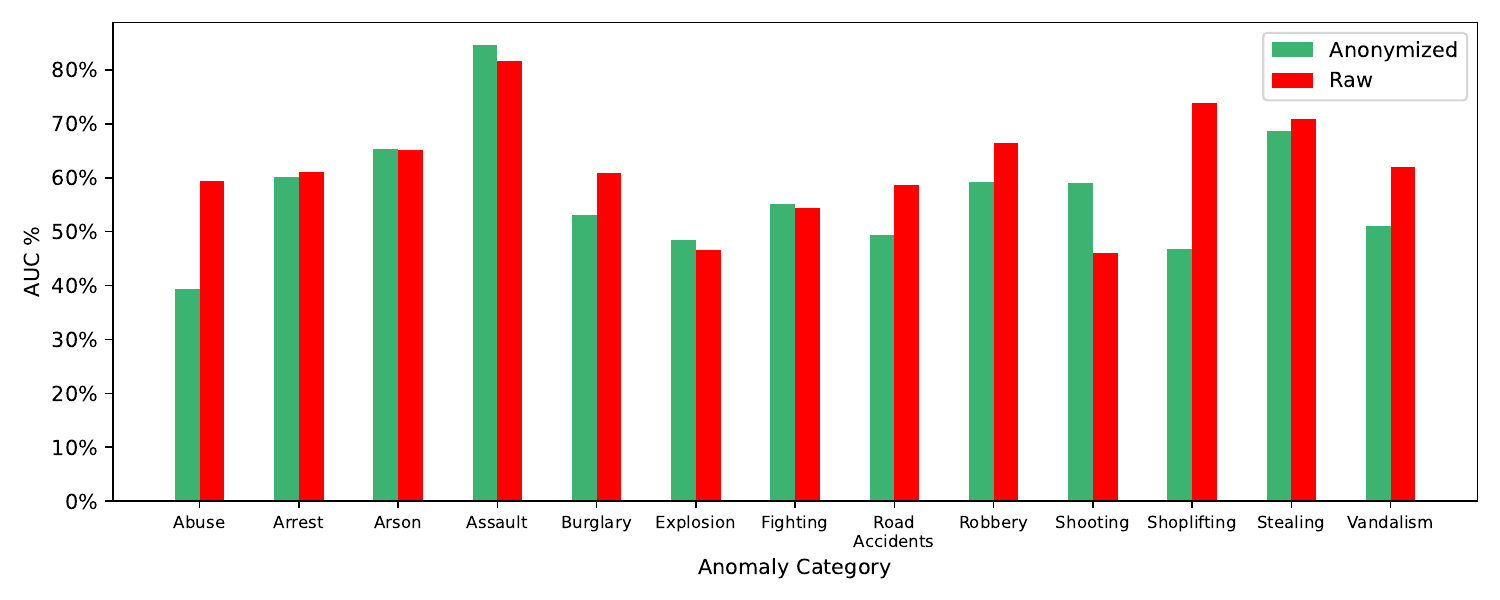}
    \vspace{-5mm}
    \caption{UCF-Crime classwise AUC performance comparison between raw and anonymized videos.}
    \label{fig:crimes_classwise}
\end{figure*}
\begin{figure*}
    \centering
    \includegraphics[width=0.75\textwidth]{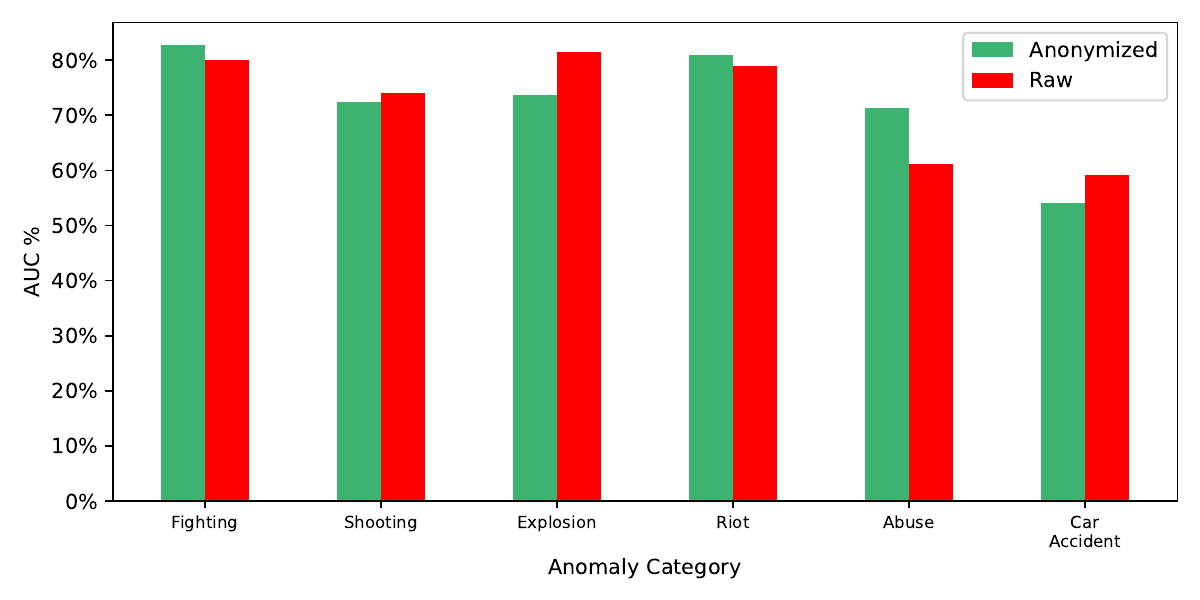}
    \vspace{-5mm}
    \caption{XD-Violence classwise AUC performance comparison between raw and anonymized videos.}
    \label{fig:crimes_classwise2}
\end{figure*}

\begin{figure*}
    \centering
    \includegraphics[width=0.99\textwidth]{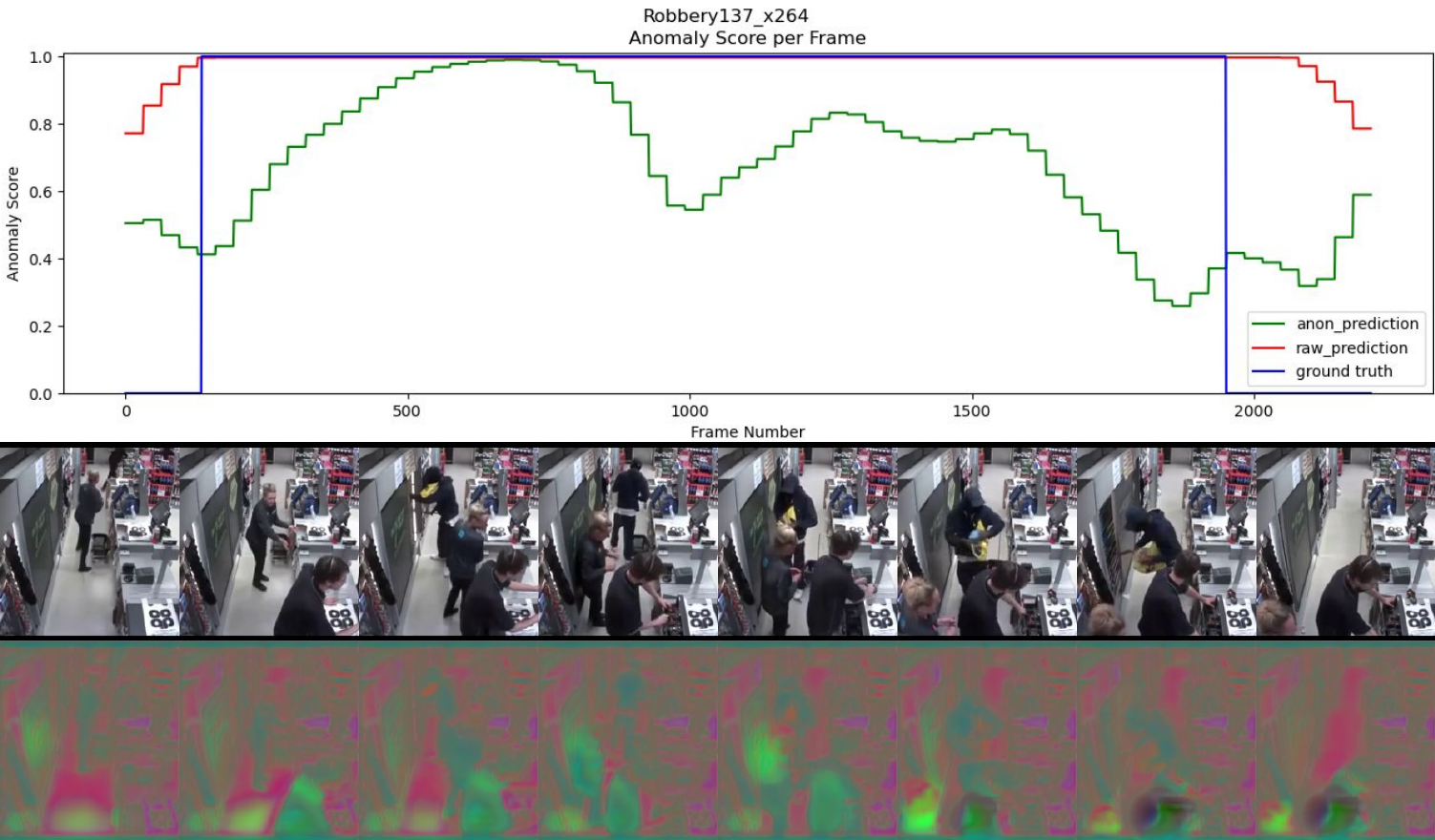}
    \caption{Frame-level anomaly score plot for video \texttt{Robbery137\_x264.mp4} from UCF-Crime. \textcolor{ForestGreen}{Green} line shows our anonymized model, \textcolor{red}{red} line is the raw input model, both compared to the \textcolor{blue}{blue} ground truth line. The below visualizations shows uniformly sampled frames from the video.}
    \label{fig:robb_plot}
\end{figure*}

\subsection{Privacy Evaluation}
To evaluate the privacy leakage of each anonymizer $f_A$, we train a ResNet50 \cite{he_deep_2015} model $f_B$ in a supervised manner to predict whether every input VISPR image contains each of the 7 private attributes from the split shown in Table~\ref{tab:vispr_split}. Training lasts for up to 100 epochs, stopping early if the learning rate drops to 1e-12. Learning rate starts at 1e-3, dropping to 1/5 of its current value on a training epoch where the loss does not decrease. Given an image $\mathbf{I^i} \in \mathbb{D}_{privacy}$, our baseline evaluates on $f_B(\mathbf{I^i})$, with subsequent experiments passing each image before evaluation, $f_B(f_A(\mathbf{I^i}))$.

\section{Results}
\label{sec:results}

\subsection{Quantitative Results}

Table~\ref{tab:main_table} compares different privacy-preserving methods and their effect on downstream anomaly detection performance. Notably, our utility loss modification allows our anonymizer to remove more privacy and improve utility performance when compared to previous methods. Compared to prior best method~\cite{spact}, our method is able to remove 19.9\% more privacy with a slightly better utility score (1.19\%).

We present class-wise performance for the anomaly detection in Fig.~\ref{fig:crimes_classwise} and ~\ref{fig:crimes_classwise2}. We also show frame-level prediction scores for the anomaly detection task in Fig. ~\ref{fig:robb_plot}.

\noindent \textbf{Effect of temporal invariance during anonymization training:}
Temporal invariance objective is conceptually opposite to temporal distinctiveness objective. With invariance, the learned representations are encouraged to be similar across the temporal dimension. Temporal invariance is implemented using the formulation from ~\cite{cvrl}. Let $\mathbf{x}^{(i)}_{t1}$ and $\mathbf{x}^{(i)}_{t'}$ be the two randomly sampled clips of a video instance $X^{(i)}$. Passing such clips through utility model $f_T$ and a non-linear projection head, we get their representations $\mathbf{z}^{(i)}_{t'}$ and $\mathbf{z}^{(i)}_{t'}$. Now the goal of the temporal invariance is to increase the mutual agreement between these two representations while maximizing the disagreement between the representation of clips of other video instances $j$, where $j\neq i$. This can be expressed as following equation:

\begin{equation}\label{eq:inv}
  \mathcal{L}_{I}=- \sum_{\substack{i=1}}^{B} \log \frac{\mathrm{h}\left(\mathbf{z}^{(i)}_{t}, \mathbf{z}^{(i)}_{t'}\right)}{\sum\limits_{j=1}^{B}[\mathbb{1}_{[j\neq i]} \mathrm{h}(\mathbf{z}^{(i)}_{t}, \mathbf{z}^{(j)}_{t}) + \mathrm{h}(\mathbf{z}^{(i)}_{t}, \mathbf{z}^{(j)}_{t'})]},
\end{equation}
\normalsize	
\noindent where $\mathrm{h}(\mathbf{u_{1}}, \mathbf{u_{2}})=\exp \left(\mathbf{u_{1}}^{T}\mathbf{u_{2}}/(\|\mathbf{u_{1}}\| \|\mathbf{u_{2}}\| \tau) \right)$ is used to compute the similarity between $\mathbf{u_{1}}$ and $\mathbf{u_{2}}$ vectors with an adjustable temperature parameter $\tau=0.1$, $B$ is batchsize. $\mathbb{1}_{[j\neq i]} \in \{0, 1\}$ is an indicator function which equals 1 iff $j \neq i$.

We perform experiments by modifying our utility loss to $\mathcal{L}_T = \mathcal{L}_{CE} + \omega*\mathcal{L}_{I}$, where $\omega$ is a loss weight.

In order to ensure that our invariance baseline is strong enough we perform several experiments varying different $\omega$ in Table~\ref{tab:ablation_invariance}. This demonstrates that temporal invariance is not well-aligned with the anomaly detection utility task. For insights, look to \main{main paper Sec. 4}.

\begin{table}[h]
\begin{center}
\begin{tabular}{l c c}
\hline

\hline

\hline\\[-3mm]
Temporal & \bf VISPR & \bf UCF-Crime \\
Invariance & Privacy & Anomaly \\
Loss Weight $\omega$ & cMAP(\%)($\downarrow$) & AUC(\%)($\uparrow$) \\
\hline
\rowcolor[rgb]{0.922,0.922,0.922} 0 & 52.71 & 73.93 \\
0.1 & 51.62 & 69.35 \\
0.5 & 46.51 & 65.84 \\
\bf 1.0 & \bf 45.64 & \bf 69.52 \\
2.0 & 52.2 & 64.4 \\
\hline

\hline

\hline\\[-3mm]
\end{tabular}
\end{center}
\vspace{-5mm}
\caption{Comparison of using different loss weights of the temporal invariance contrastive loss during the anonymization process. Bold indicates best trade-off.}

\label{tab:ablation_invariance}
\end{table}

\noindent \textbf{Effectiveness of different $f_T$ architectures:}
For all experiments in the main paper, we follow previous works and use I3D \cite{carreira_quo_2018}. 
Table~\ref{tab:ft_arch} shows experiments with different $f_T$ architectures to ensure that our anonymization function is suitable for varying architectures. Since the downstream anomaly detection task relies on input features, it is important to note that our I3D implementation outputs features of dimensionality 2048, while MViTv2 \cite{li2021improved} and R3D-18 \cite{hara3dcnns} output 768 and 512, respectively. These experiments used the same hyperparameters as our best I3D experiment, the models may achieve a better trade-off with hyperparameter tuning.

\begin{table}[h]
\begin{center}
\begin{tabular}{l c c}
\hline

\hline

\hline\\[-3mm]
$f_T$ & \bf VISPR & \bf UCF-Crime \\
Model & Privacy & Anomaly \\
Architecture $\lambda$ & cMAP(\%)($\downarrow$) & AUC(\%)($\uparrow$) \\
\hline
I3D & 42.21 & 74.81 \\
MViTv2 & 24.21 & 69.22 \\
R3D-18 & 33.58 & 70.67 \\
\hline

\hline

\hline\\[-3mm]
\end{tabular}
\end{center}
\vspace{-6mm}
\caption{Comparison of different $f_T$ architectures for both the proxy utility task and feature extraction.}

\label{tab:ft_arch}
\end{table}

\subsection{Qualitative Results}
We present qualitative results of our anonymization function in Fig.~\ref{fig:xd_viz} and ~\ref{fig:st_viz}.
More visualization can be found in the attached videos of the supplementary material.

\begin{figure*}
    \centering
    \includegraphics[width=0.95\textwidth]{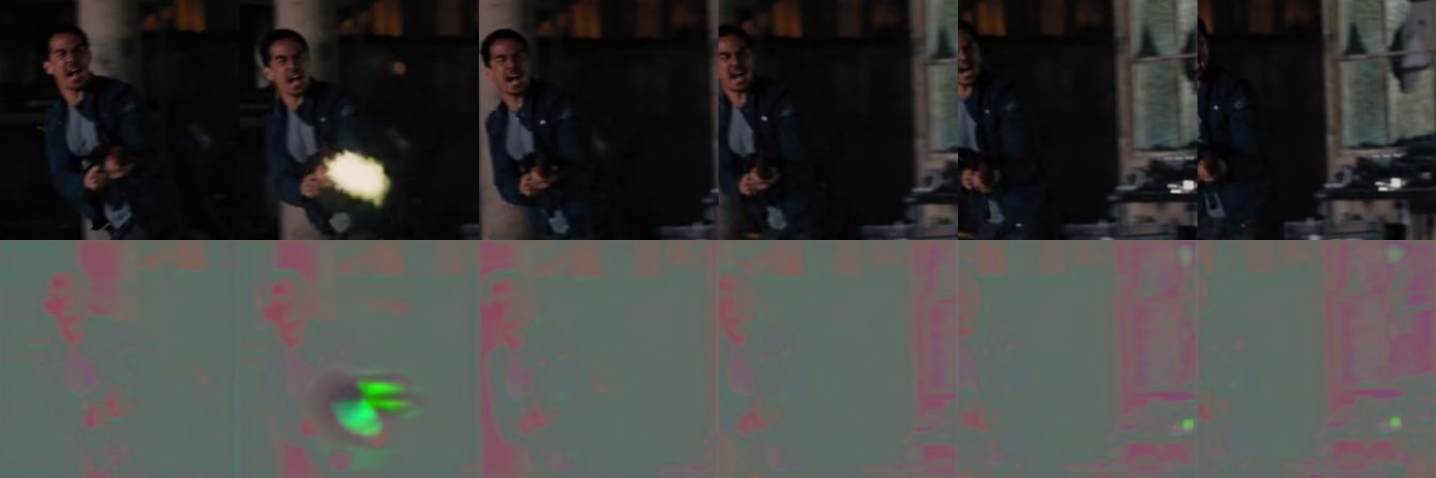}
    \captionsetup{justification=centering}
    \caption{Visualization of anomalous clip (shooting) from XD-Violence dataset video\\ \texttt{Fast.Furious.6.2013\_\_\#00-45-40\_00-47-13\_label\_B2-0-0.mp4}.}
    \label{fig:xd_viz}
\end{figure*}

\begin{figure*}
    \centering
    \includegraphics[width=0.95\textwidth]{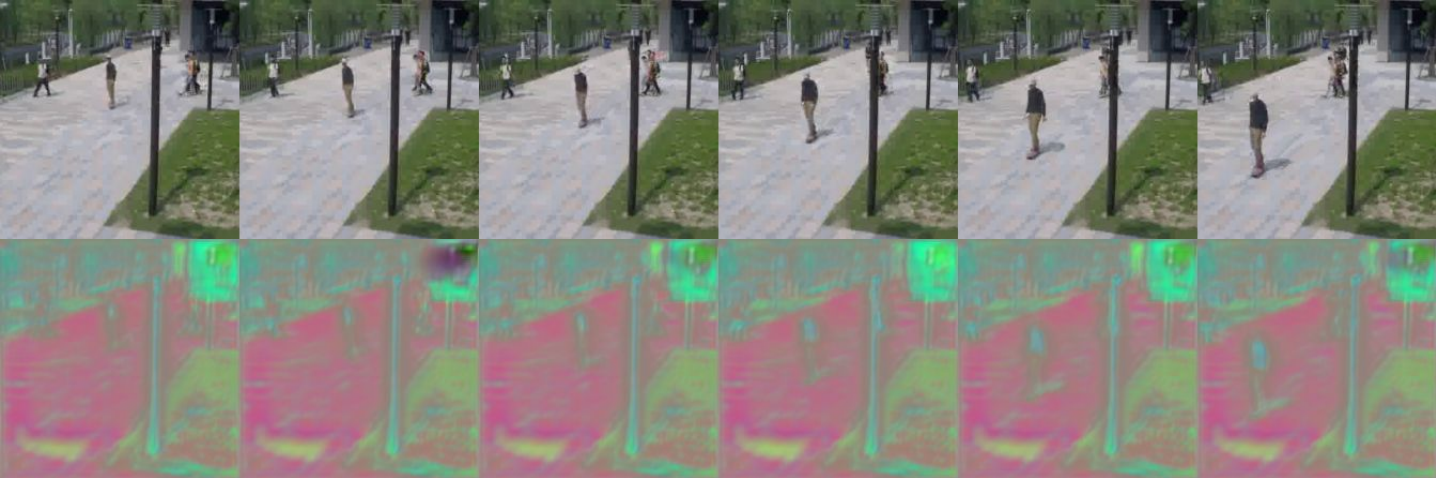}
    \caption{Visualization of anomalous clip (skateboard passing) from ShanghaiTech dataset video \texttt{08\_0178.avi}.}
    \label{fig:st_viz}
\end{figure*}

\subsection{Training Progression}
\begin{figure*}[h]
    \includegraphics[width=\textwidth]{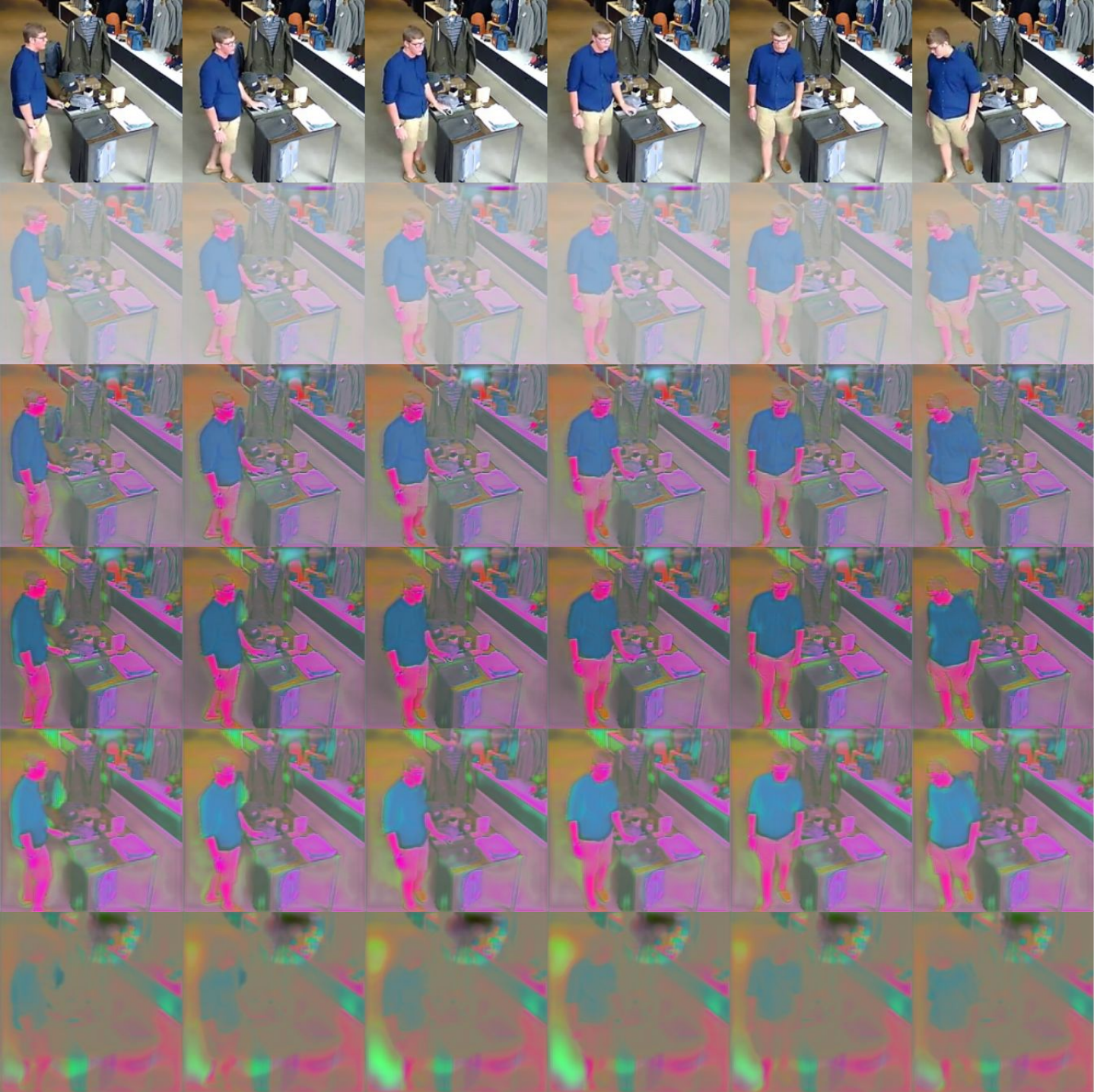}
    \caption{Training progression per epoch of the anonymization process. In order from top to bottom, visualization after $f_A$ on epoch 1, 6, 9, 12, 15, and 20 is shown.}
    \label{fig:epoch_prog}
\end{figure*}

\begin{figure*}[h]
    \includegraphics[width=\textwidth]{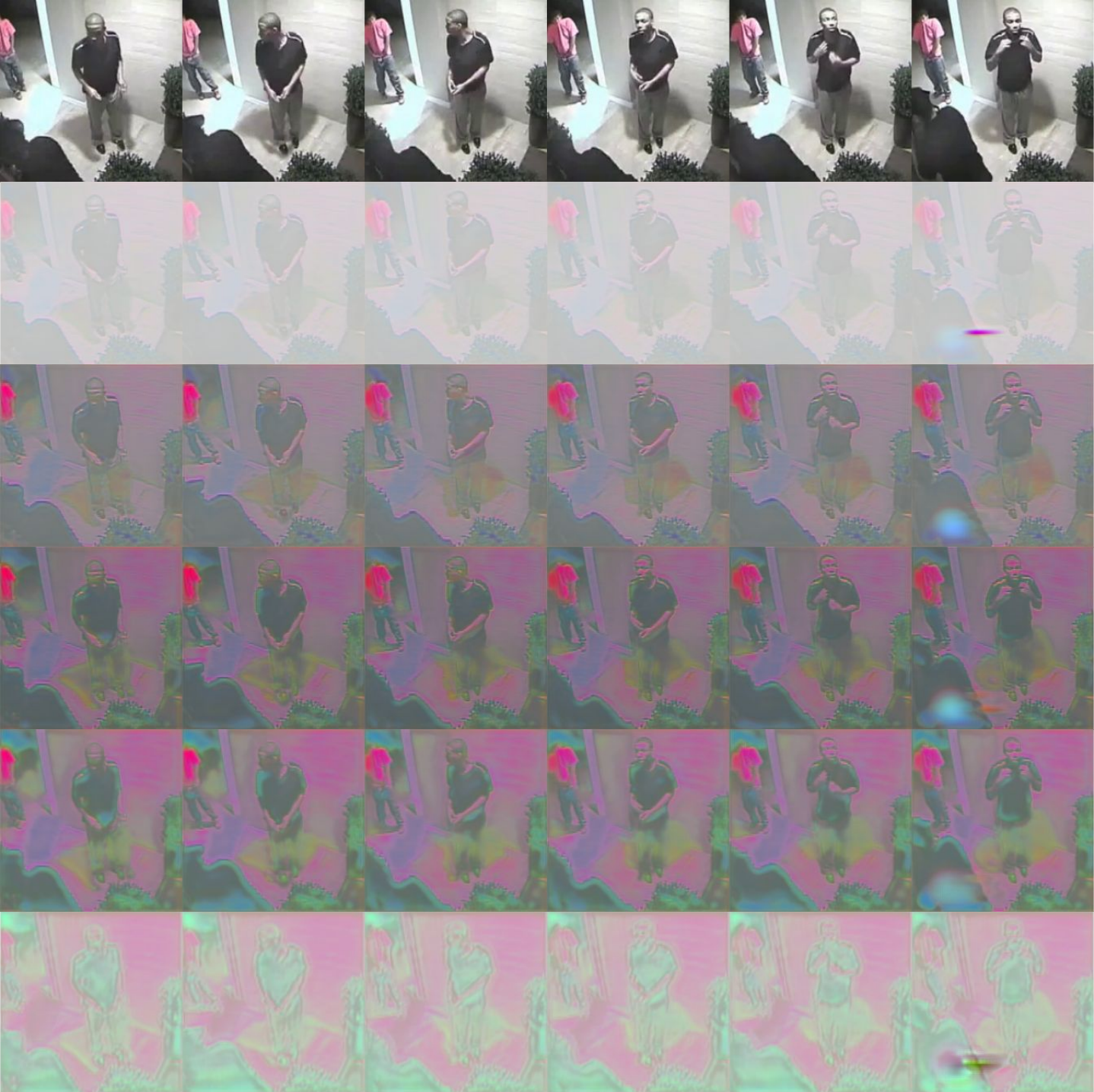}
    \caption{Training progression per epoch of the anonymization process. In order from top to bottom, visualization after $f_A$ on epoch 1, 6, 9, 12, 15, and 20 is shown.}
    \label{fig:epoch_prog2}
\end{figure*}

We show outputs of our anonymization framework at different epochs of anonymization training in Fig.~\ref{fig:epoch_prog} and ~\ref{fig:epoch_prog2}. We can clearly observe that as the training progresses, our framework is able to anonymize better.

\newpage
\mbox{~}
\clearpage

\end{document}